\documentclass[twocolumn]{article}
\usepackage[utf8]{inputenc}

\usepackage{graphicx}

\usepackage{amsmath}
\usepackage{nccmath}
\usepackage{algorithm}
\usepackage{algpseudocode}
\usepackage{colortbl}
\usepackage{amsfonts}
\usepackage[numbers]{natbib}
\usepackage{balance}
\usepackage{comment}

\usepackage{ragged2e}
\usepackage{subfig}
\usepackage[final]{microtype}
\usepackage{fancyhdr}

\newcommand{\changed}[1]{\textcolor[rgb]{0.8,0.0,0.4}{{}{#1}}}

\fancypagestyle{firstpage}
{
    \fancyhf{}

    \fancyhead[L]{\small \justifying \noindent \textbf{Cite as: Cruz, F., Dazeley, R., Vamplew, P., Moreira, I. (2021). Explainable robotic systems: understanding goal-driven actions in a reinforcement learning scenario. Neural Computing and Applications.} 
    }

    \fancyfoot[L]{\small \justifying \noindent This preprint has not undergone peer review or any post-submission improvements or corrections. The Version of Record of this article is published in Neural Computing and Applications, and is available online at https://doi.org/10.1007/s00521-021-06425-5.
    }
    
}

\title{Explainable robotic systems: Understanding goal-driven actions in a reinforcement learning scenario}
\author{Francisco Cruz$^{1,3}$ \and
Richard Dazeley$^1$ \and
Peter Vamplew$^2$ \and
Ithan Moreira$^3$}
\date{\normalsize
    $^1$ School of Information Technology, Deakin University, Geelong, Australia.\\
    $^2$ School of Science, Engineering and Information Technology, Federation University, Ballarat, Australia.\\
    $^3$ Escuela de Ingenier\'ia, Universidad Central de Chile, Santiago, Chile.\\
    Corresponding e-mails: francisco.cruz@deakin.edu.au, richard.dazeley@deakin.edu.au, p.vamplew@federation.edu.au, 
    ithan.moreira@alumnos.ucentral.cl\\
}

\begin{document}

\maketitle

\begin{abstract}
Robotic systems are more present in our society everyday.
In human-robot environments, it is crucial that end-users may correctly understand their robotic team-partners, in order to collaboratively complete a task.
To increase action understanding, users demand more explainability about the  decisions by the robot in particular situations.
Recently, explainable robotic systems have emerged as an alternative focused not only on completing  a task satisfactorily, but also on justifying, in a human-like manner, the reasons that lead to making a decision.
In reinforcement learning scenarios, a great effort has been focused on providing explanations using data-driven approaches, particularly from the visual input modality in deep learning-based systems.
In this work, we focus rather on the decision-making process of reinforcement learning agents performing a task in a robotic scenario.
Experimental results are obtained using 3 different set-ups, namely, a deterministic navigation task, a stochastic navigation task, and a continuous visual-based sorting object task.
As a way to explain the goal-driven robot's actions, we use the probability of success computed by three different proposed approaches: memory-based, learning-based, and introspection-based.
The difference between these approaches is the amount of memory required to compute or estimate the probability of success as well as the kind of reinforcement learning representation where they could be used.
In this regard, we use the memory-based approach as a baseline since it is obtained directly from the agent's observations.
When comparing the learning-based and the introspection-based approaches to this baseline, both are found to be suitable alternatives to compute the probability of success, obtaining high levels of similarity when compared using both the Pearson's correlation and the mean squared error.
\end{abstract}

\textbf{Keywords:} Explainable robotic systems, Explainable reinforcement learning, Goal-driven explanations, Human-robot environment, Human-aligned artificial intelligence.

\thispagestyle{firstpage}

\section{Introduction}
Explainable robotic systems have become an interesting field of research since they give robots the ability to explain their behavior to the human counterpart~\cite{Anjomshoae19, Sheh17}. 
One of the main benefits of explainability is to increase the trust in Human-Robot Interaction (HRI) scenarios~\cite{Wang15, Wang16}.

To get robots into our daily-life environments, an explainable robotic system should provide clear explanations specially focused on non-expert end-users in order to understand the robot's decisions~\cite{rosenfeld2019explainability}.
Often, such explanations, given by robotic systems, have been focused on interpreting the agent's decision based on its perception of the environment~\cite{Pocius19, Lengerich17}. 
Whereas, little work has provided explanations of how the action selected is expected to help it to achieve its goal~\cite{sado2020explainable}. 
For instance, the use of the visual sensory modality often attempts to understand how a deep neural network makes decisions considering a visual state representation.
In general, prior work has been focused on state-based explanations~\cite{Hendricks16, Li18, Iyer18}, i.e., given explanations take the form of: \textit{`I chose action $A$ because of this feature $F$ of the state'}.
Nevertheless, explanations from a goal-oriented perspective have so far been less addressed and, therefore, there exists a gap between explaining the robot's behavior from state features and the help to achieve its aims~\cite{Anjomshoae19}.

In this work, we propose two robotic scenarios where the robot has to learn a task using reinforcement learning (RL)~\cite{Sutton18}.
The aim of RL is to provide an autonomous agent with the ability to learn a new skill by interacting with the environment.
While RL has been shown to be an effective learning approach in diverse areas such as human cognition~\cite{Gershman17, Palminteri17}, developmental robotics~\cite{cruz2018multi, cruz2018improving, churamani2020icub}, and games~\cite{Kempka16, Vinyals17, barros2020learning}, among others, 
an open issue is the lack of a mechanism that allows the agent to clearly communicate the reasons why it chooses certain actions given particular goals~\cite{sado2020explainable}.
Therefore, it is not easy for a person to entrust important tasks to an AI-based system (e.g. robots) that cannot justify its reasoning~\cite{Adadi18}. 
In this regard, our goal-driven approach is concerned with the outcomes of each decision, e.g., an explanation may take the form of: \textit{`action $A_1$ gives an 85\% chance of success the task compared to 38\% for action $A_2$'}.

When interacting with the environment, an RL agent will learn a policy to decide which action to take from a certain state.
In value-based RL methods the policy uses Q-values to determine the course of actions to follow. 
The Q-values represent the value function in reinforcement learning.
The value function is an estimation of possible future collected reward starting from an specific situation and following an specific policy.
Therefore in practise, the Q-values used to choose what action to perform next.
However, the Q-values are not necessarily meaningful in the problem domain, but rather in the reward function domain~\cite{juozapaitis2019explainable}.
Hence, they do not allow the robot to explain its behavior in a simple manner to a person with no knowledge about RL or machine learning techniques.
It is clearly not acceptable for the agent to provide explanations to non-expert end-users such as: \textit{`I choose action $A_1$ because it maximizes future collected reward'} or \textit{`I choose action $A_2$ because it is the next one following the optimal policy'}~\cite{degraff2017people}.

In this paper, we propose three different approaches that allow a learning agent to explain, using a user-friendly concept, the decision of selecting an action over the other possible ones.
In these approaches, explanations are given using the probability of success, i.e., the probability of accomplishing the task following particular criteria related to the scenario.
Although we are not automatically generating explanations yet, using the probability of success an RL agent might next explain its behavior not only in terms of Q-values or the probability of selecting an action, but rather in terms of the necessity to complete the intended task.
For instance, using a Q-value, an agent might provide an explanation as: \textit{`I decided to go left because the Q-value associated with the action left in the current state is $-0.181$'}, which could be the highest Q-value in the reward function domain but is pointless for a non-expert user.
Whereas using the probability of success, an agent might give an explanation as: \textit{`I chose to go left because that has a $73.6\%$ probability of reaching the goal successfully'}, which is much more understandable for a non-expert end-user.

The proposed methods differ in their approach to determining the probability of success. 
Each approach trades off accuracy against space complexity (understood as the memory required for each method) or applicable problem domain. 
This paper illustrates empirically the benefits and shortcomings of each approach. 
We have tested our approaches in 3 simulated robotic scenarios. 
The first 2 scenarios consists of a robot navigation task where we have used both deterministic and stochastic state transitions. 
The third scenario consists of a sorting object task where we have used continuous visual inputs in order to test how the introspection-based approach scales up to more complex scenarios.

The first method uses a memory-based explainable RL approach, which we have previously introduced~\cite{cruz2019memory} to compute the probability of success in both bounded and unbounded grid-world scenarios.
This approach uses a high space complexity, but provides a realistic probability of success since it is obtained directly from the actual agent’s observations. 
The memory-based approach is used as a baseline to verify the accuracy of the other two estimation methods developed in this paper. 
The new approaches in this paper include a learning-based and an introspection-based method.
We hypothesize they significantly reduce the space complexity while obtaining similar results. 
This reduction in space complexity would also allow these methods to be better suited for domains requiring a continuous state representation.
The learning-based method allows the agent to learn the probability of success while the Q-values are learned.
The introspection-based method uses a logarithmic transformation to compute the probability of success directly from the Q-values.


\begin{figure*}
  \centering
  \includegraphics[width=0.7\linewidth]{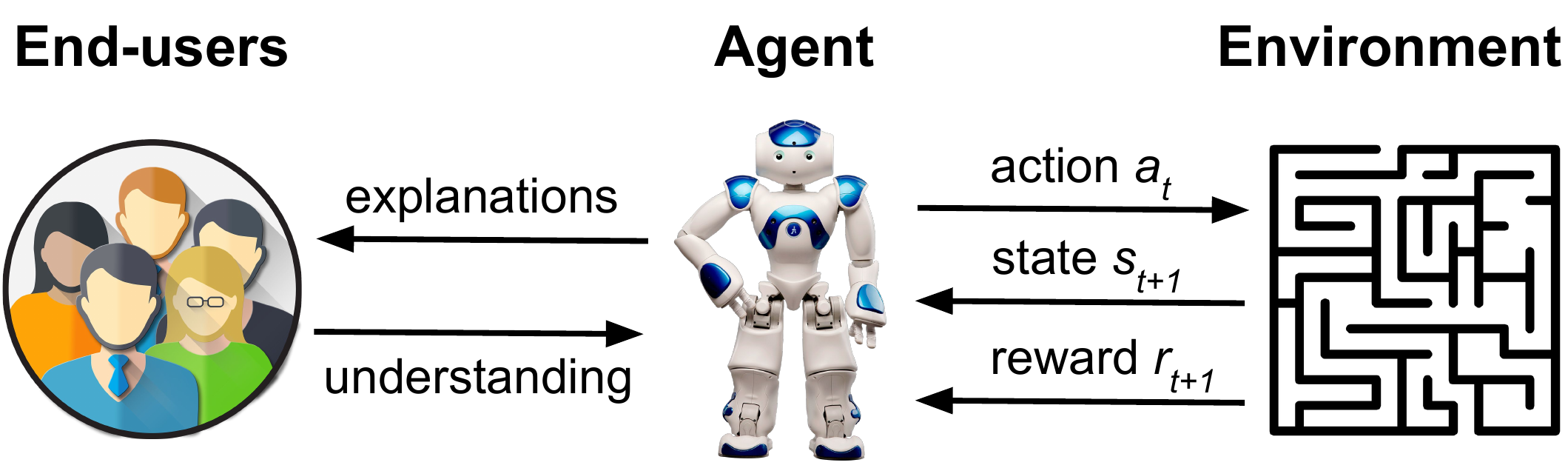}
  \caption{The explainable reinforcement learning framework. 
  The environment's response is a new state $s_{t+1}$ and sometimes a reward $r_{t+1}$.
  During and after the learning process, the agent is able to explain its behavior to the end-users, which leads to an increased level of action understanding.
  }
  \label{fig:ExplainableReinforcementLearning}
\end{figure*}

\section{Explainable reinforcement\\learning in robotics}

\subsection{Reinforcement learning}


RL is studied as a decision-making mechanism in both cognitive and artificial agents~\cite{Cangelosi15}.
Unlike other learning techniques, RL does not include supervision nor instruction but rather the agent is able to sense changes in the environment due to carried out actions.
Hence, new states and numeric rewards are observed by the agent every time-step after performing an action. 
In other words, an agent must learn from its own experience~\cite{Kober13}.
The usual interactive loop between an RL agent and its environment is depicted on the right-hand side of Fig.~\ref{fig:ExplainableReinforcementLearning}, in the interaction between the agent and the environment, where it is possible to observe that the robot performs an action $a_t$ from the state $s_t$ to reach a new state $s_{t+1}$ and a reward signal $r_{t+1}$ as a response from the environment.

An RL agent must learn a policy $\pi: S \rightarrow A$, with $S$ being a set of states and $A$ a set of actions available from $S$.
By learning a policy $\pi$, the agent looks for the highest possible cumulative future reward starting from a state $s_t$~\cite{Sutton18}.
The optimal policy and the optimal action-value function are referred to as $\pi^*$ and $q^*$ respectively.


The~ optimal~ action-value~ function~ is~ defined~ as 
$q^{*}(s,a) = \underset{\pi} {\mathrm{max}} ~ q^{\pi}(s,a)$ 
and solved using the Bellman optimality equation for $q^*$, as shown in Eq. \eqref{Eq:Bellman}.


\begin{eqnarray}
q^{*}(s_t,a_t) & = \sum_{s_{t+1}} & p(s_{t+1}|s_t,a_t) [ r(s_t,a_t,s_{t+1}) + \nonumber\\
& &\gamma \max_{a_{t+1}} q^{*} (s_{t+1},a_{t+1}) ]
\label{Eq:Bellman}
\end{eqnarray} 
where $s_t$ represents the agent's current state, $a_t$ the performed action, $r$ is the reward signal received by the agent after performing $a_t$ from $s_t$ to reach the state $s_{t+1}$, and $a_{t+1}$ is a possible action from $s_{t+1}$.
Moreover, $p$ is the probability of reaching $s_{t+1}$ given the agent's current state $s_t$ and the taken action $a_t$.

To solve Eq.~\ref{Eq:Bellman}, an alternative is to use the on-policy method SARSA~\cite{Rummery94}.
SARSA is a temporal-difference learning method which iteratively updates the state-action values $Q(s,a)$ using the Eq. \eqref{Eq:SARSARL} as follows:


\begin{eqnarray}
Q(s_t,a_t)&\leftarrow&Q(s_t,a_t)+\alpha[r_{t+1}+\gamma Q(s_{t+1},a_{t+1})- \nonumber\\
& &Q(s_t,a_t)]
\label{Eq:SARSARL}
\end{eqnarray} 

Thus, the optimal action-value function maps states to actions to maximize the future reward ($r$) over a time ($t$) as shown in Eq. (\ref{Eq:Action-Value}). 
In the equation, $\mathbb{E}_{\pi}[ ]$ denotes the expected value given that the agent follows policy $\pi$.
In continuous representations, an approximation function (e.g., implemented by deep neural networks) allows an agent to work with high-dimensional observation spaces, such as pixels of an image~\cite{mnih2015human}.

\begin{eqnarray}
    Q^{*}(s_t,a_t) & = & max ( \mathbb{E}_{\pi}[r_{t}+\gamma r_{t+1}+\gamma^{2}  r_{t+2}+... \nonumber \\ 
    & & | s_t=s,a_t=a,\pi] )
    \label{Eq:Action-Value}
\end{eqnarray}

%
%

\subsection{Explainable reinforcement\\learning}

Machine learning techniques are getting more attention everyday in different areas of our daily life.
Applications in fields such as robotics, autonomous driving cars, assistive companions, video games, among others are constantly shown in the media~\cite{Gunning17}.
There are different alternatives to model intelligent agents, e.g., by using phenomenological (white-box) models, empirical (black-box) models, or hybrid (grey-box) models~\cite{cruz2007indirect, cruz2010indirect}. 
Explainable artificial intelligence (XAI) has emerged as a prominent research area that aims to provide black-box AI-based systems the ability to give human-like and user-friendly explanations to non-expert end-users~\cite{Miller18, dazeley2021levels}.
XAI research is motivated by the need to provide transparent decision-making that people can trust and accept~\cite{Fox17}.

Up to now, there has been considerable effort to provide agents with instruments to explain their actions to end-users.
Nevertheless, in explainable reinforcement learning (XRL), most works have addressed explainability using technical explanations~\cite{puiutta2020explainable}.
For instance,~\cite{Verma18} presented the programmatically interpretable reinforcement learning framework which works only with symbolic inputs considering deterministic policies, but not covering stochastic policies that are commonly pre-sent in real-world scenarios. 
Another approach~\cite{Shu17} included hierarchical and interpretable skill acquisition. 
In this approach, the tasks are decomposed in a hierarchical plan with understandable actions.
Moreover,~\cite{Hein18} introduced a hybrid approach for interpretable policies mixing RL with genetic programming.
The approach explains the policies using equations instead of more human-friendly explanations.
In~\cite{juozapaitis2019explainable} the drQ method was presented, an XRL approach based on reward decomposition~\cite{erwig2018explaining}.
Using a variant of Q-learning, the method decomposed the action-values to better understand the agent's action selection.
However, the proposed approach focused on RL practitioners and does not look for end-user explanations.

Additionally, in~\cite{Pocius19} saliency maps are used to explain the agent's decisions.
In this work, a game scenario is used to provide visual explanations with deep RL.
In~\cite{Wang18} an explainable recommendation system is presented within an RL scheme.
Inspired by cognitive science,~\cite{Madumal19} introduced a game scenario providing causal explanations determined by causal models. 
However, the particular causal model has to be known in advance for the specific domain.
Recently,~\cite{madumal2020distal} presented a distal explanation model analyzing opportunity chains and counterfactuals by using decision trees and causal models. 
As in the previous case, the main drawback is the need for a previously known causal model which is not possible in large-domain problems.
In~\cite{dazeley2021conceptual} a conceptual framework for XRL is introduced based on the causal explanation network model representing explanations of people's behavior~\cite{bohm2015people}.
The proposed casual XRL framework includes a simplified version using the more relevant components for zero-order explanations~\cite{dazeley2021levels}.
While surveying the main XRL approaches, this work also brings insights about the future development of the XRL field, such as the explanation of goals.

Another closely related field is explainable planning, whose primary goal is to help end-users to better understand the plan produced by a planner~\cite{Fox17}.
In~\cite{Sukkerd18}, Sukkerd et al. propose a multi-objective probabilistic planner for a simple robot navigation task.
They provide verbal explanations of quality-attribute objectives and~ properties,~ however,~ these~ rely~ on~ assumptions about the preference structure on quality attributes.

\subsection{Explainable robotic systems}

As robots are giving their first steps into domestic scenarios, they are more likely to work with humans in teams.
Therefore, if a robot has the ability to explain its behavior to non-expert users, this may lead to an increase in the human's understanding of the robot's actions~\cite{anderson2019explaining} as shown within Fig.~\ref{fig:ExplainableReinforcementLearning}.
In this regard, some works have demonstrated explanations to be an effective way of increasing understanding and trust in HRI scenarios.
For instance, Want et al.~\cite{Wang16} proposed a domain-independent approach to generate explanations and measure their impact in trust with behavioral data from a simulated human-robot team task.
Their experiments showed that using explanations improved transparency, trust and team performance.
Lomas et al.~\cite{Lomas12} developed a prototype system to answer people's questions about robot actions.
They assumed the robot uses a cost map-based planner for low-level actions and a finite state machine high-level planner to respond to specific questions.
Yang et al.~\cite{Yang17} presented a 
dual-task environment, where they treated the trust as an evolving variable, in terms of the user experience.

Furthermore, Sander et al.~\cite{Sanders14} proposed to use different modalities to evaluate the effect on transparency in an HRI scenario.
They also varied the level of information provided by the robot to the human and measured the understanding responses.
In their study, participants reported higher understanding levels when the level of information was constant, however, no significant differences were found when using a different communication modality.
Haspiel et al.~\cite{Haspiel18} carried out a study about the importance of timing when giving explanations. 
They used four different automated vehicle driving~ conditions,~ namely,~ no~ explanation,~ explanation seven seconds before an action, one second before an action, and seven seconds after an action.
They found that earlier explanations lead to higher understanding by end-users.

In HRI environments, explainable agency~\cite{Langley16, Anjomshoae19} has been used to refer to robots explaining the reasons or motivations of the underlying decision-making process.
For instance, in~\cite{Tabrez19} an HRI environment was presented where users informed that the robot was more intelligent, helpful, and useful when giving explanations about its behavior.
In~\cite{Sequeira19a} an explainability framework was presented using a three-level analysis from the agent's transition history.
Furthermore, an extension included a study of the agent's capabilities and limitations~\cite{Sequeira19b}.
Finally,~\cite{Langley17} proposed an episodic memory to save state transitions during the learning process, however, the approach was not implemented.

\section{Goal-driven explanations}
As discussed previously, the actions taken by a robot may be explained using technical terms, for instance, comparing Q-values or algorithmically speaking.
However, this is not possible to do it successfully if the explanation is addressed to people with no technical knowledge about algorithms or machine learning~\cite{juozapaitis2019explainable}.
In this regard, this work looks for explanations that can be understood by any person interacting with a robot, using terminology similar to human interaction. 

On several occasions, intelligent agents and robots deployed in domestic-like environments have to interact with non-expert end-users.
An important current research challenge is to improve the ability a robot has to communicate explanations about intentions and performed actions to interacting users~\cite{Gunning17}, especially in case of failure~\cite{Dulac19}.

As aforementioned, although there is increasing literature in different XAI subfields, such as explainable planners, interpretable RL, or explainable agency, just a few works are addressing the XRL challenge in robotic scenarios.
In some of those works, although they are in a certain way focused on XRL, they have different aims than ours, e.g., to explain the learning process using saliency maps from a computational vision perspective, especially when using deep reinforcement learning as in~\cite{Pocius19, greydanus2018visualizing}.
In this paper, we focus on explaining goal-oriented decisions to provide an understanding to the user of what motivates the robot's specific actions from different states, taking into account the problem domain.

In HRI scenarios, there are many questions which could arise from a non-expert user when interacting with a robot. Such questions include \textit{what}, \textit{why}, \textit{why not}, \textit{what if}, \textit{how to}~\cite{Lim09}. 
%
%
However, from a non-expert end-user perspective, we can consider the most relevant questions as to 'why?' and 'why not?'~\cite{Madumal19}, e.g., 'why did the agent perform action $a$ from state $s$'?
Hence, we focus this approach on answering these kinds of questions using an understandable domain language. 
Thus, our approach explains how the agent's selected action is the preferred choice based on its likelihood of achieving its goal. 
This is achieved by determining the probability of success. 
Once the probability of reaching the final state is determined the agent will be able to provide the end-user an  understandable explanation of why one action is preferred over others when in a particular state.

In the following subsections, we present different approaches aiming to explain why an agent selects an action in a specific situation given an specific goal.
As discussed, we focus our analysis on the probability of success as a way to support the agent's decision as this will be more intuitive for a non-expert user than an explanation based directly on the Q-values.
We introduce three different approaches to estimate the probability of success. 
The memory-based approach develops a transition network of the domain during learning and has been previously presented by us 
applied to a grid-world scenario~\cite{cruz2019memory}; 
whereas the learning-based approach uses a so-called $\mathbb{P}$-value learned in parallel with the Q-value. 
The introspection-based approach proposes a model of self-examination to relate the agent's own motivations and actions directly to the probability of success using a numerical transformation of the Q-values. 
Therefore, in this work, we extend our previous approach adding two additional alternatives to compute the probability of success with the aim to use fewer memory resources and to be usable in non-deterministic, continuous, and deep RL domains.

\subsection{Memory-based approach}

In~\cite{cruz2019memory}, we proposed a memory-based explainable reinforcement learning approach to compute the probability of success $P_s$ using an RL agent with an episodic memory as suggested in~\cite{Langley17}.
The episodic memory was implemented using a list ($T_{List}$) to save all the state-action pairs transited by the agent through the learning process. 

Using the total number of transitions $T_t$ and the number of transitions in a success sequence $T_s$, we compute the probability of success $P_s$.
In order to obtain $T_t$ and $T_s$, the transitions saved in the episodic memory are used, represented as state-action pairs within the list $T_{List}$.
Therefore, when the robot reaches the goal position, the path followed determines the probability $P_s \leftarrow T_s/T_t$. 

The proposed algorithm implements the on-policy method SARSA~\cite{Rummery94} with softmax action selection.
Algorithm~\ref{Alg:XRLMemory} shows the memory-based approach to train the robot using the episodic memory. 
Particularly, line~\ref{Alg:Lin:Save} saves into the episodic memory the transited state-action pairs, and in line~\ref{Alg:Lin:ComputePs} the final probabilities of success $P_s$ is computed after each learning episode.

\begin{algorithm}[t]
\begin{algorithmic}[1]
\State Initialize $Q(s,a)$, $T_t$, $T_s$, $P_s$
\For{each episode}
\State Initialize $T_{List}[]$, $s_t$
\State Choose an action $a_t$ from $s_t$
\Repeat 
\State Take action $a_t$
\State Save state-action transition $T_{List}$.add($s_t$, $a_t$) \label{Alg:Lin:Save}
\State $T_t[s_t][a_t] \leftarrow T_t[s_t][a_t] + 1$
\State Observe reward $r_{t+1}$ and next state $s_{t+1}$
\State Choose next action $a_{t+1}$ using softmax action \textcolor{white}{...... ..............} selection method
\State $Q(s_t,a_t)\leftarrow Q(s_t,a_t)+\alpha[r_{t+1}+\gamma Q(s_{t+1},a_{t+1})$\textcolor{white}{. ............................}$ - Q(s_t,a_t)]$
\State $s_t\leftarrow s_{t+1}$; $a_t\leftarrow a_{t+1}$
\Until {$s_t$ is terminal (goal or aversive state)}
\If {$s_t$ is goal state}
\For{each s,a $\in T_{List}$}\
\State $T_s[s][a] \leftarrow T_s[s][a] + 1$
\EndFor
\EndIf
\State Compute $P_s \leftarrow T_s / T_t$ \label{Alg:Lin:ComputePs}
\EndFor
\end{algorithmic}
\caption{Explainable reinforcement learning approach to compute the probability of success using the memory-based approach.} 
\label{Alg:XRLMemory}
\end{algorithm}

It has been previously shown that an agent using this memory-based approach is able to explain its behavior using the probability of success in an understandable manner for non-expert end-users at any moment during its operation~\cite{cruz2019memory}.
However, in this approach, the use of memory increases rapidly and the use of resources is $\sim O(s \times a \times l \times n)$, where $s$ is the number of states, $a$ is the number of actions per state, $l$ the average length of the episodes, and $n$ the number of episodes.
Therefore, this approach is not suitable to high/continuous state or action situations, such as real-world robotics scenarios.

\subsection{Learning-based approach}
Although~ to~ compute~ the~ probability~ of success by means of the episodic memory is possible and has previously lead to good results, one of the main problems is the increasing amount of memory needed as the problem dimensionality enlarges.
In this regard, an alternative to explain the behavior in terms of the probability of success is by learning it through the agent's learning process.

To learn the probability of success, we propose to maintain an additional set of state-action values. 
Therefore, learning in parallel the probability of success using a state-action transition table is also an alternative to explain the behavior to non-expert users. 
We refer to this table as a $\mathbb{P}$-table and as $\mathbb{P}$-value to an individual value inside the table.
While using the memory-based approach the episodic memory could increase unrestricted, when using a $\mathbb{P}$-table, as proposed in this learning-based approach, the additional memory needed is fixed to the size of the $\mathbb{P}$-table, i.e., $\sim O(s \times a)$.
This represents a doubling of the memory requirements of RL which implies a constant increase to the base algorithm and is therefore negligible.

Similarly to Q-values, to learn the $\mathbb {P}$-values implies to update the estimations after each performed action.
In our approach, we employ the same learning rate $\alpha$ for both values, however, the main change with respect to the implemented temporal-difference algorithm is that we do not use a reduced discount factor $\gamma$, or in other words, we set it to $\gamma = 1$ to consider the total sum of future rewards.
From the discount reward perspective, to use $\gamma = 1$ does not represent an issue to solve the underlying optimization problem since the Q-values are used for learning purposes, which indeed use a discount factor $\gamma$ to guarantee the convergence.

As using the discount factor $\gamma=1$, the agent associates each action based on the total sum of all future rewards, nevertheless, we want to learn a $\mathbb {P}$-value as a probability of success $\in [0,1]$.
Therefore, we do not use the reward $r_{t+1}$ to update the $\mathbb {P}$-table, instead, we use a success flag $\varphi_{t+1}$ which consists of a value equal to $0$ to indicate that the task is being failed, or a value equal to $1$ to indicate that the task has been completed.
In such a way, the agent learns the probability of success considering the sum of the probabilities to finalize the task in the future.

The update of the $\mathbb{P}$-values is performed according to Eq. \eqref{Eq:learning} as follows:

\begin{eqnarray}
\mathbb{P}(s_t,a_t) & \leftarrow & \mathbb{P}(s_t,a_t) + \alpha [ \varphi_{t+1} + \mathbb{P}(s_{t+1},a_{t+1}) - \nonumber\\
 & & \mathbb{P}(s_t,a_t) ]
\label{Eq:learning}
\end{eqnarray} 
where $a_t$ is the taken action at the state $s_t$. 
$\mathbb{P}(s_t,a_t)$ and $\mathbb{P}(s_{t+1},a_{t+1})$ are the probability of success values considering the state and the action at timestep $t$ and $t+1$ respectively. 
Moreover, $\alpha$ is the learning rate and $\varphi_{t+1}$ is the success flag used to indicate if the task has been completed or not.
Note that $\gamma$ is not present in Eq. \eqref{Eq:learning} since its value is set to 1. 

Algorithm~\ref{Alg:XRLLearning} shows the learning-based approach.
Particularly, in line~\ref{Alg:Lin:UpdateP} the $\mathbb{P}$-table is updated. 
While the $\mathbb{P}$-table and Q-table share a similar structure and learning mechanism, their roles are quite different with the Q-values driving the agent's policy while the $\mathbb{P}$-values are used for explanatory purposes. 
Separating the role of these values allows the use of $\varphi_{t}$ and $r_{t}$ terms best suited to each purpose.
Therefore, $\varphi_{t}$ might be sparse with non-zero values only when the task is successfully completed, whereas $r_{t}$ might be free to take on any form, including incorporation of reward shaping terms~\cite{Ng99} which might be difficult to interpret for a non-expert.

\begin{algorithm}[t]
\begin{algorithmic}[1]
\State Initialize $Q(s,a)$, $\mathbb{P}(s_t,a_t)$
\For{each episode}
\State Initialize $s_t$
\State Choose an action $a_t$ from $s_t$
\Repeat 
\State Take action $a_t$
\State Observe reward $r_{t+1}$ and next state $s_{t+1}$
\State Choose next action $a_{t+1}$ using softmax action \textcolor{white}{...... ..............} selection method
\State $Q(s_t,a_t)\leftarrow Q(s_t,a_t)+\alpha[r_{t+1}+\gamma Q(s_{t+1},a_{t+1})$\textcolor{white}{. ............................}$ - Q(s_t,a_t)]$
\State $\mathbb{P}(s_t,a_t) \leftarrow \mathbb{P}(s_t,a_t) + \alpha [ \varphi_{t+1} + \mathbb{P}(s_{t+1},a_{t+1})$\textcolor{white}{. .............................}$ - \mathbb{P}(s_t,a_t) ]$ \label{Alg:Lin:UpdateP}
\State $s_t\leftarrow s_{t+1}$; $a_t\leftarrow a_{t+1}$
\Until {$s_t$ is terminal (goal or aversive state)}
\EndFor
\end{algorithmic}
\caption{Explainable reinforcement learning approach to compute the probability of success using the learning-based approach.} 
\label{Alg:XRLLearning}
\end{algorithm}

The learning method described in Eq. \eqref{Eq:learning} suits only for discrete representations.
However, the same learning-based approach can be extended to continuous and larger discrete scenarios by using a function approximators, such as neural networks.
As usual, we would need a neural approximator to learn the Q-values and an additional neural network to learn the $\mathbb{P}$-values in parallel.

\subsection{Introspection-based approach}
Even though the learning-based approach previously presented represents an improvement when compared to the memory-based approach, in terms of using fewer memory resources, it still requires some additional memory to keep the $\mathbb{P}$-values updated.
Moreover, during the learning process, time is also needed for computations and for learning episodes in order to obtain a better estimation.

Therefore, we also look for an approach that allows us to estimate the probability of success $\hat{P_s}$ directly from the Q-values using a numerical transformation.
As pointed out, the idea of this approach is to relate the Q-values to the probability of success as an introspective means of the agent's self-motivation. 
In this regard, this approach is more efficient in terms of used memory and time required for the learning, i.e., since it is computed directly from the Q-values, it does not require additional memory leading to the use of resources of $\sim O(1)$.

Bearing in mind the temporal difference learning approach shown in Eq. \eqref{Eq:SARSARL}, the optimal Q-values represent possible future reward, therefore, they are expressed in the reward function domain.
Thus, if an agent reaches a terminal state in an episodic task obtaining a reward $R^T$, the associated Q-value approximates this reward.
In a simplified manner, we can consider any Q-value $Q(s,a)$ as the terminal reward $R^T$ multiplied by the times the discount factor $\gamma$ is applied, as 
$Q(s,a) \approx R^T \cdot \gamma ^ n$.
%
%
Using this derivation, when $Q(s,a)$ converges to the true value, we can compute how far away the agent is from obtaining the total reward for any state, using directly the Q-values.
Therefore, using the previous argument, we have computed the estimated distance $n$ to the reward, as shown in Eq. \eqref{Eq:n}:

\begin{ceqn}
\begin{align}
n \approx log_\gamma \frac{Q(s,a)}{R^T}
\label{Eq:n}
\end{align}
\end{ceqn}
where $Q(s,a)$ is the Q-value, $R^T$ the reward obtained when the task is completed successfully, and $n$ is the estimated distance, in number of actions, to the reward.

After computing the estimated distance $n$ to obtain a reward, we use this value as the base to estimate an estimated probability of success $\hat{P}_s$.
Using a constant transformation, we weight the estimated distance $n$ by $\frac{1}{2 \cdot log_{\gamma} 10} + 1$.
Therefore, what we are actually performing is a logarithmic base transformation to estimate the probability of success $\hat{P}_s$ as shown in Eq.~\eqref{Eq:nWeight} and Eq.~\eqref{Eq:nWeightLog10}. 
In our approach, we also take into account stochastic transitions represented by the $\sigma$ parameter.
We will discuss this parameter further in the following section.

\begin{ceqn}
\begin{align}
\hat{P}_s \approx (1 - \sigma) \cdot \left( \frac{n}{2 \cdot log_{\gamma}10} + 1 \right)
\label{Eq:nWeight}
\end{align}
\end{ceqn}

\begin{ceqn}
\begin{align}
\hat{P}_s \approx (1 - \sigma) \cdot \left( \frac{1}{2} \cdot log_{10} \frac{Q(s,a)}{R^T} + 1 \right)
\label{Eq:nWeightLog10}
\end{align}
\end{ceqn}

This transformation is carried out in order to shape the probability of success curve as a common base logarithm (base 10) that fits the behavior of both previously introduced approaches.
Moreover, we shift the curve to a region where the probability values are plausible by adding $1$ and multiplying by $1/2$. 
Finally, in order to restrict the value of the probability of success, considering $\hat{P}_s \in [0,1]$, we compute the rectification shown in Eq. \eqref{Eq:PsBound}, which basically consists of assigning a value of $0$ when the result is less than that, or $1$ when the result is greater than $1$.

\begin{ceqn}
\begin{align}
\hat{P}_s \approx \left[ (1 - \sigma) \cdot \left( \frac{1}{2} \cdot log_{10} \frac{Q^*(s,a)}{R^T} + 1 \right) \right]^{\hat{P}_s \le 1}_{\hat{P}_s \ge 0}
\label{Eq:PsBound}
\end{align}
\end{ceqn}

In algorithm~\ref{Alg:XRLIntrospection}, the estimated probability of success $\hat{P}_s$ is computed at the end in line~\ref{Alg:Lin:PFromQ}.
Although algorithms~\ref{Alg:XRLMemory},~\ref{Alg:XRLLearning}, and~\ref{Alg:XRLIntrospection} share a common structure based on the temporal-difference learning, we decided to show them independently since they do differ in the way they compute the probability of success inside the SARSA method. 
The three approaches introduced in this section are tested experimentally in a simulated robotic set-up described in the following section.


\begin{algorithm}[t]
\begin{algorithmic}[1]
\State Initialize $Q(s,a)$, $\hat{P}_s$
\For{each episode}
\State Initialize $s_t$
\State Choose an action $a_t$ from $s_t$
\Repeat 
\State Take action $a_t$
\State Observe reward $r_{t+1}$ and next state $s_{t+1}$
\State Choose next action $a_{t+1}$ using softmax action \textcolor{white}{...... ..............} selection method
\State $Q(s_t,a_t)\leftarrow Q(s_t,a_t)+\alpha[r_{t+1}+\gamma Q(s_{t+1},a_{t+1})$\textcolor{white}{. ............................}$ - Q(s_t,a_t)]$
\State $s_t\leftarrow s_{t+1}$; $a_t\leftarrow a_{t+1}$
\Until {$s_t$ is terminal (goal or aversive state)}
\State $\hat{P}_s \approx \left[ (1 - \sigma) \cdot \left( \frac{1}{2} \cdot log_{10} \frac{Q(s_t,a_t)}{R^T} + 1 \right) \right]^{\hat{P}_s \le 1}_{\hat{P}_s \ge 0}$ \label{Alg:Lin:PFromQ}
\EndFor
\end{algorithmic}
\caption{Explainable reinforcement learning approach to compute the probability of success using the introspection-based approach.} 
\label{Alg:XRLIntrospection}
\end{algorithm}

\section{Experimental set-ups}
In this section, we describe the experimental scenarios used to test the proposed approaches.
We have designed 3 domestic scenarios using the CoppeliaSim robot simulator~\cite{Rohmer13}.
The first 2 experiments use a discrete robot navigation task including deterministic and stochastic transitions.
In this scenario, a humanoid robot needs to reach a goal position by moving across different rooms from an initial position.
The third experiment is a continuous visual-based sorting task.
In this scenario, a robotic arm has to sort different objects with different shape and color using raw images from an RGB camera as inputs.

\begin{figure*}
  \centering
  \includegraphics[width=1.0\linewidth]{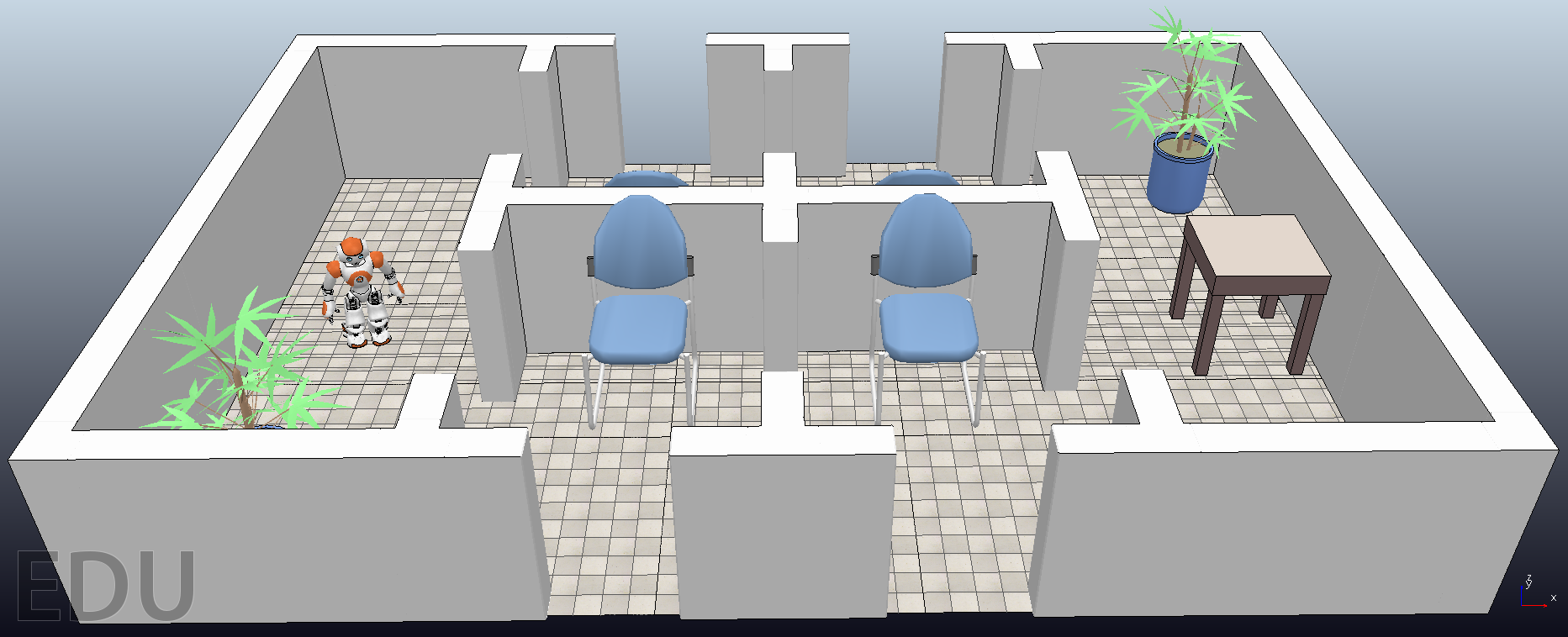}
  \caption{The simulated robot navigation task. The initial position is in the room to the left, the so-called room 0. The goal is to reach the table in room 5 to the right. 
  From the initial position, two different paths are possible to reach the goal. In the intermediate rooms, namely rooms 1, 2, 3, and 4, the agent may leave the level, and if so, it cannot get back and needs to restart a new learning episode (thereby failing to complete the task).}
  \label{fig:RobotNavigationTask}
\end{figure*}

\subsection{Robot navigation task} \label{sec:RobotNavigationTask}

The first scenario used in this work is an episodic robot task in which the transitions can be modeled as a graph, and therefore the Q-values can be approximated using a tabular method.
The scenario consists of a simulated robot navigation task comprising of six rooms and three possible actions to perform from each room.
The simulated scenario is shown in Fig.~\ref{fig:RobotNavigationTask}.
Furthermore, two variations are considered for the proposed scenario: deterministic and stochastic transitions.

In the proposed scenario, a mobile robot has to learn how to navigate from a fixed initial position (room 0) through different rooms considering two possible paths to find the table within the goal position (room 5).
Moreover, it is also possible to observe that every room in the middle of the paths, i.e., from room $1$ to room $4$, has an exit that leaves the level.
These transitions are treated as leading to an aversive region and, therefore, once any of these exits have been taken, the robot is unable to come back and needs to stop the current learning episode and restart a new one from the initial position.
In the scenario, the agent starts from the initial position and may transit two symmetric paths towards the goal position.

We have defined six states corresponding to the six rooms and three possible actions from each state.
Actions are defined taking into account the robot's perspective.
The possible actions are 
(i) $a_L$, move through the left door, (ii) $a_R$, move through the right door, and (iii) $a_S$, stay in the same room.
These transitions are, in principle, all deterministic. 
Nevertheless, in some situations, transitions may not be deterministic and may include a certain level of stochasticity or uncertainty, as in partially observable problems, or also sometimes due to noisy sensors, for instance.
Therefore, taking into consideration these situations, we have also used a parameter $\sigma \in [0, 1]$ to include stochastic transitions.

When stochastic transitions are used, the next state reached, as a result of performing an action, is any of the possible future states from where the action has been performed. 
For instance, if the action $a_L$ is being performed from the state $s_0$, the agent is expected to be in the state $s_1$ once the action is completed, taking into account deterministic transitions ($\sigma=0$).
However, considering stochasticity (i.e., $\sigma>0$), there is also a probability the agent could finish in the state $s_0$ or the state $s_2$ since these two are also reachable from the state $s_0$ (by performing the action $a_S$ and the action $a_R$ respectively).

For the learning process, the reward function returns a positive reward of $1$ when the agent reaches the final state and a negative reward of $-1$ in case the agent reaches an aversive region, i.e., when it leaves the scenario. Eq.~\ref{Eq:reward} shows the reward function for $s$.

\begin{equation}
r(s) = \left\{
\begin{array}{r l}
  1 & \textrm{if } s \textrm{ is the final state}\\
 -1 & \textrm{if } s \textrm{ is an aversive state}\\
\end{array}
\right.
\label{Eq:reward}
\end{equation}

Although we are aware that this scenario is rather simple from the learning perspective, and thus manageable by an RL agent, in this work we are focusing on giving a basis for goal-oriented explanations regarding fundamental RL aspects using the proposed methods. 
In this regard, this robot navigation task allows us to obtain valuable initial insights about the behavior of the different proposed approaches and their equivalence.
Moreover, following we propose a second scenario in which later we test the approach consuming less resources in order to prove the scalability to more complex scenarios.

\subsection{Sorting object task}

The second scenario used in this work is a continuous visual-based sorting object task~\cite{moreira2020deep}.
An alternative in continuous states is to approximate the agent's state directly from raw images.
For instance, deep RL uses the same RL structure adding a deep neural network in order to approximate the Q-values from raw inputs.

\begin{figure}[t]
  \centering
  \includegraphics[width=1\linewidth]{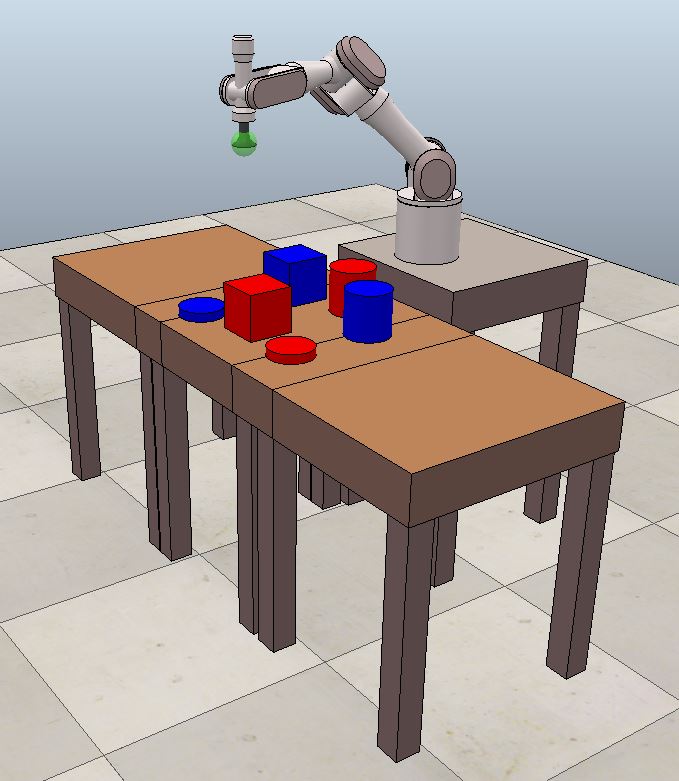}
  \caption{The simulated sorting object task. The objects are initially located on the central table. The robot arm must grab each object and sort them in the lateral tables. The task is completed once the robot has sorted out all the objects. Moreover, if an object is wrongly sorted the learning episode finishes and the task is restarted.
  }
  \label{fig:SortingObjectTask}
\end{figure}

In this scenario, a robotic arm has to sort 6 different objects.
The objects are represented by geometric figures with different colors.
The sorting task involves moving the objects from a central table to two other tables located to each side, one for each class of object. 
Initially, the objects are placed in the central table on random positions.
Figure~\ref{fig:SortingObjectTask} illustrates the sorting object task scenario.

The robot is allowed to perform four different actions. 
The possible actions are 
(i) grab an object, (ii) drop an object, (iii) move right, and (iv) move left.


To grab and drop an object, the robot has a suction pad that is activated or deactivated respectively.
Move right or left, it moves the robot's arm to the indicated position from the robot's perspective.
When moving, the suction pad can either have an object or be free.
The actions to perform are decided by a neural network by approximating the Q-values, while all the low-level control commands to produce the robot movements are computed using inverse kinematics.

The robot's state is represented with raw images containing $64 \times 64$ pixels obtained directly from an RGB camera.
The raw image is presented to the neural network previously normalized as $\in[0,1]$.

The reward function is defined also considering subgoals, i.e., the robot receives a reward for each object correctly sorted.
Moreover, when all the objects are properly sorted, the task is completed and an additional reward is received.
If an object is wrongly sorted, the training episode finishes and a negative reward is obtained.
Finally, after 18 learning time-steps (the minimal number of steps to complete the task), the robot receives an additional small negative reward in order to encourage it to finish the task as soon as possible.
The reward function is shown following:

\begin{equation}
    r(s) = \left\{
    \begin{array}{r l}
          1 & \textrm{if all the objects are sorted}\\
         0.4 & \textrm{if a single object is sorted}\\
          -1 & \textrm{if an object is incorrectly sorted}\\
         -0.01 & \textrm{if steps } > 18\\
    \end{array}
    \right.
    \label{Eq:reward2}
\end{equation}

Thus, if the agent correctly finalizes the task in 18 steps, it receives a total reward of 3, i.e., 0.4 for each of the first 5 objects sorted, and an additional reward of 1 due to the sixth and final object.

\begin{figure}[t]
  \centering
  \includegraphics[width=1\linewidth]{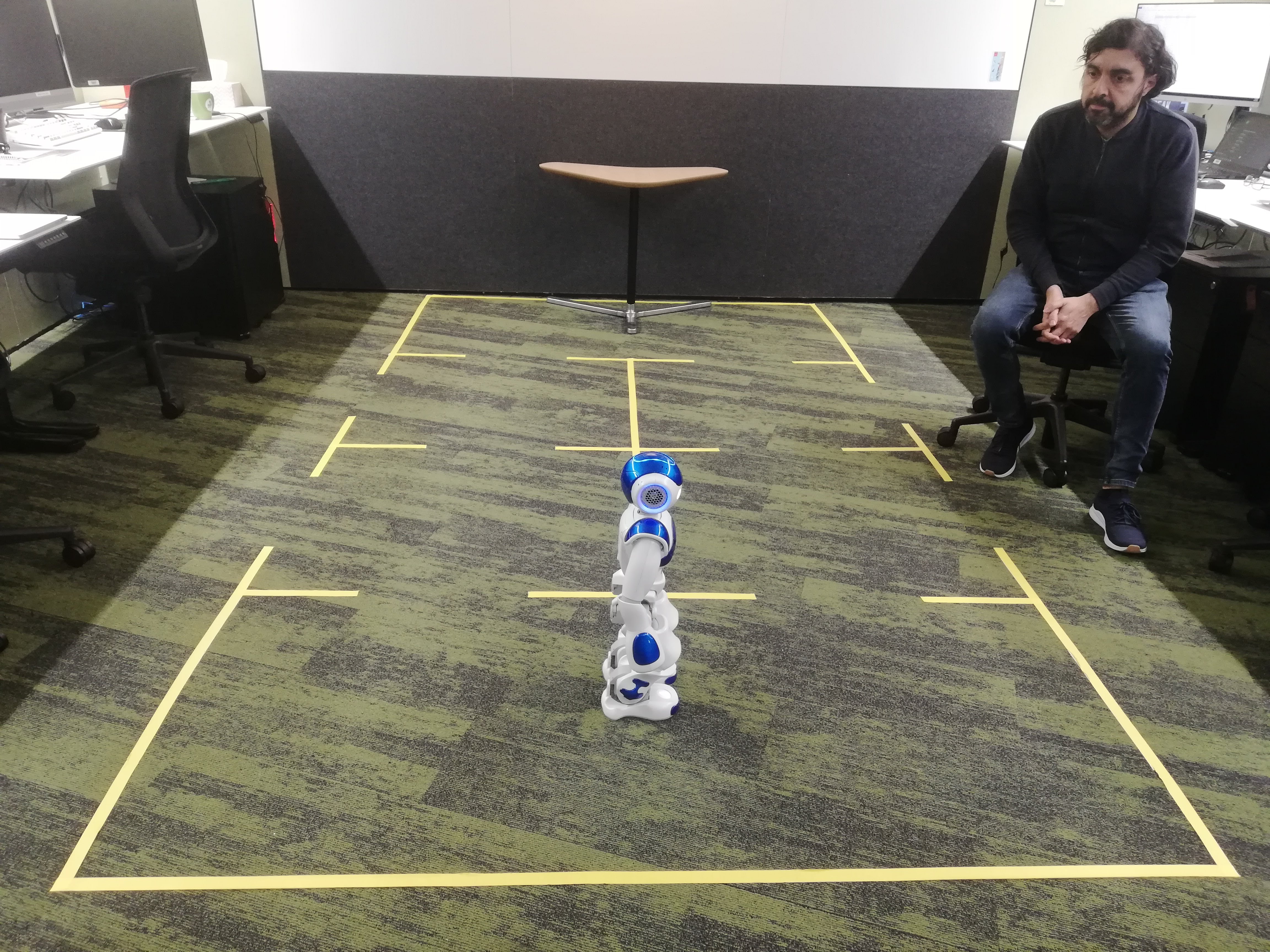}
  \caption{The robot navigation task in a real-world scenario. 
  In this setup, the robot interacts with an end-user. 
  The robot provides different kinds of explanations using either the probability of success or a Q-value. 
  End-users can evaluate each received explanation in terms of usefulness to explain the robot's behavior.
  }
  \label{fig:RealScenario}
\end{figure}

\subsection{Real-world proof of concept} \label{sec:RealScenario}

Although this work has been primarily developed in simulated scenarios, we have also run a simple real-world test in order to assess the application with end-users.
We have replicated the robot navigation task described in Section \ref{sec:RobotNavigationTask} in a real-world scenario and we have tested our approach with an end-user.
The replicated real-world scenario is shown in Fig. \ref{fig:RealScenario}.

Using this scenario, we have carried out 2 different situations.
In the first situation, the robot performed an action from the initial state (room 0).
In the second situation, the robot performed an action that led it to the goal position (room 5).
In both situations, the end-user received an explanation built from either the probability of success or the Q-values.

\section{Experimental results}

In this section, we describe the results obtained when using the proposed explainable approaches in the scenarios described in the last section.
During the robot navigation task, the experiments have been performed using the on-policy learning algorithm SARSA, as shown in Eq.~\ref{Eq:SARSARL}, and the softmax action selection method, where $a = \dfrac{e^{Q(s_t,a)/\tau}}{\sum_{a_i \in A} e^{Q(s_t,a_i)/\tau}}$, where $s_t$ is the current agent's state, $a_i$ an action in the set $A$ of available actions, $\tau$ is the temperature parameter, and $e$ the exponential function.
A total of $20$ agents have been trained and thus, the plots in this work depict average results.
The parameters used for the training are: learning rate $\alpha=0.3$, discount factor $\gamma=0.9$, and softmax temperature $\tau=0.25$.

During the sorting object task, a continuous visual representation is used for the agent's state.
Therefore, a function approximator for $Q(s_t,a_t)$ (as shown in Eq.~(\ref{Eq:Action-Value})) and experience replay~\cite{adam2012experience} are introduced for generalization during the learning process.
We implemented the Deep Q-learning algorithm using previous experiences to train a convolutional neural network (CNN) as an approximator for Q-values.
The CNN architecture comprises an input of $64 \times 64$ pixels in RGB channels, 3 convolutional layers ($8 \times 8$ with 4 filters, $4 \times 4$ with 8 filter, and $2 \times 2$ with 16 filters, respectively). 
Each convolutional layer is followed by a $2 \times 2$ max-pooling layer.
Finally, there is a flattened layer and a fully connected layer with 256 neurons.
The output of the network includes a softmax function with 4 neurons to represent the Q-value of each possible action. 
The experience replay technique uses a memory $M$ containing 128 tuples that include $<s_t, a_t, r_t, s_{t+1}>$.
Initially, 1000 random actions are performed as a pretraining in order to populate the memory $M$.
Afterwards, the agent learning is carried out.
In this experiment, due to the higher learning complexity, only $10$ agents have been trained, 
however as mentioned, plots still show average results.
The parameters used for the training are: $\epsilon$-greedy action selection with initial $\epsilon=1$ and $\epsilon$ decay rate = 0.9995, learning rate $\alpha=0.001$, and discount factor $\gamma=0.9$.
Although the training parameters used in these experiments are not much relevant in terms of explainability, we include them for reference.
Certainly, they do affect the agent's learning speed, nevertheless, we focus on better understanding and explaining the agent's decisions.

\subsection{Deterministic robot navigation\\task}

Initially, we have tested an RL agent moving across the rooms considering deterministic transitions, i.e., performing an action $a$ from the state $s_i$ to state $s_j$ always reaching the intended state $s_j$ with probability equal to $1$, or $\sigma = 0$. 
For the analysis, following we plot the obtained Q-values, estimated distance $n$ to task completion, and probabilities of success using the three proposed methods.
In principle, we show these values in all cases for the actions performed from the initial state $s_0$, nevertheless, similar plots can be obtained for each state.

\begin{figure*}[!hbt]
\centering
\subfloat[Q-values.]{\includegraphics[width=0.33\textwidth]{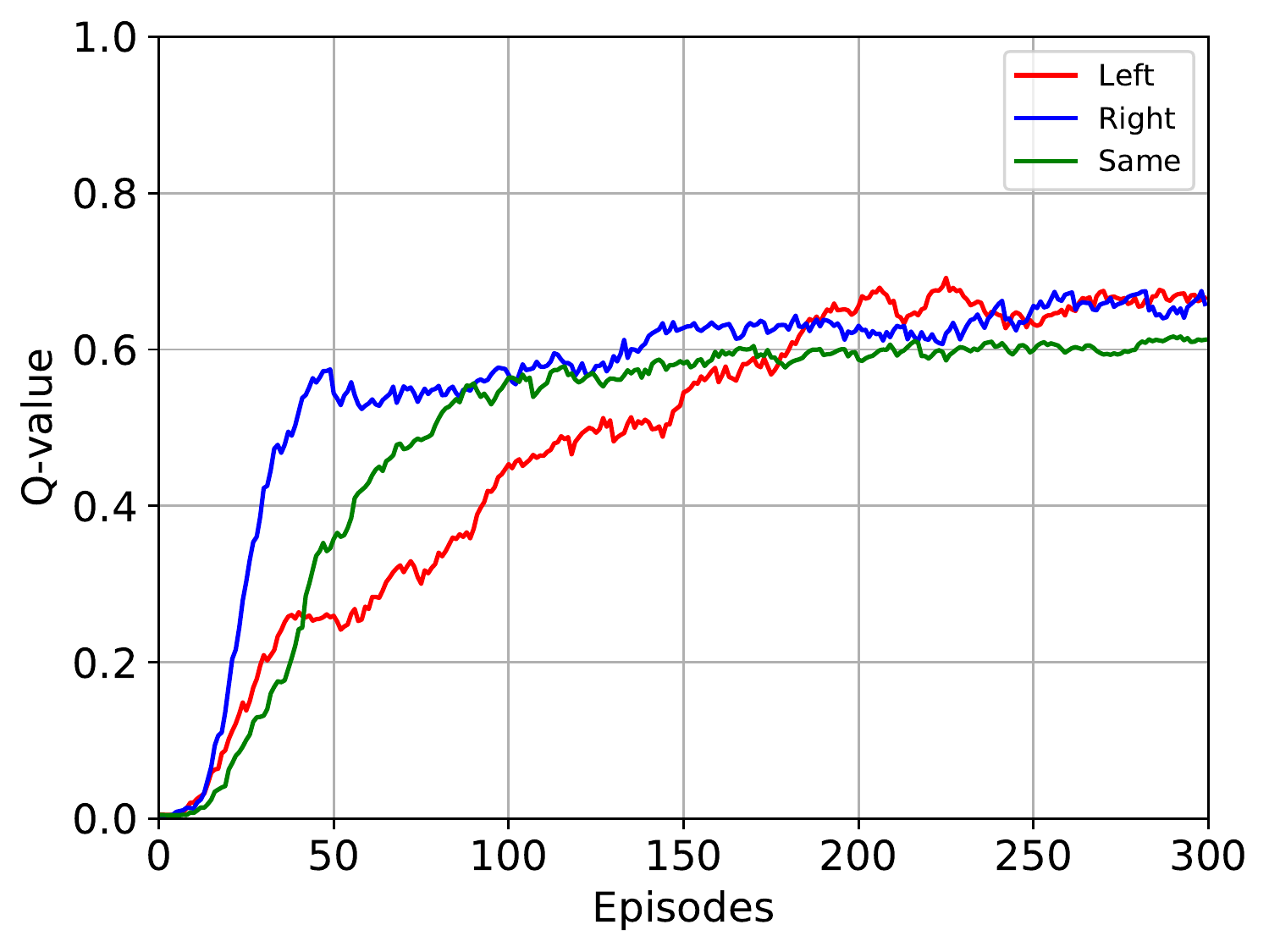} \label{fig:QValuesDeterministic}} 
\subfloat[Estimated distance n.]{\includegraphics[width=0.33\textwidth]{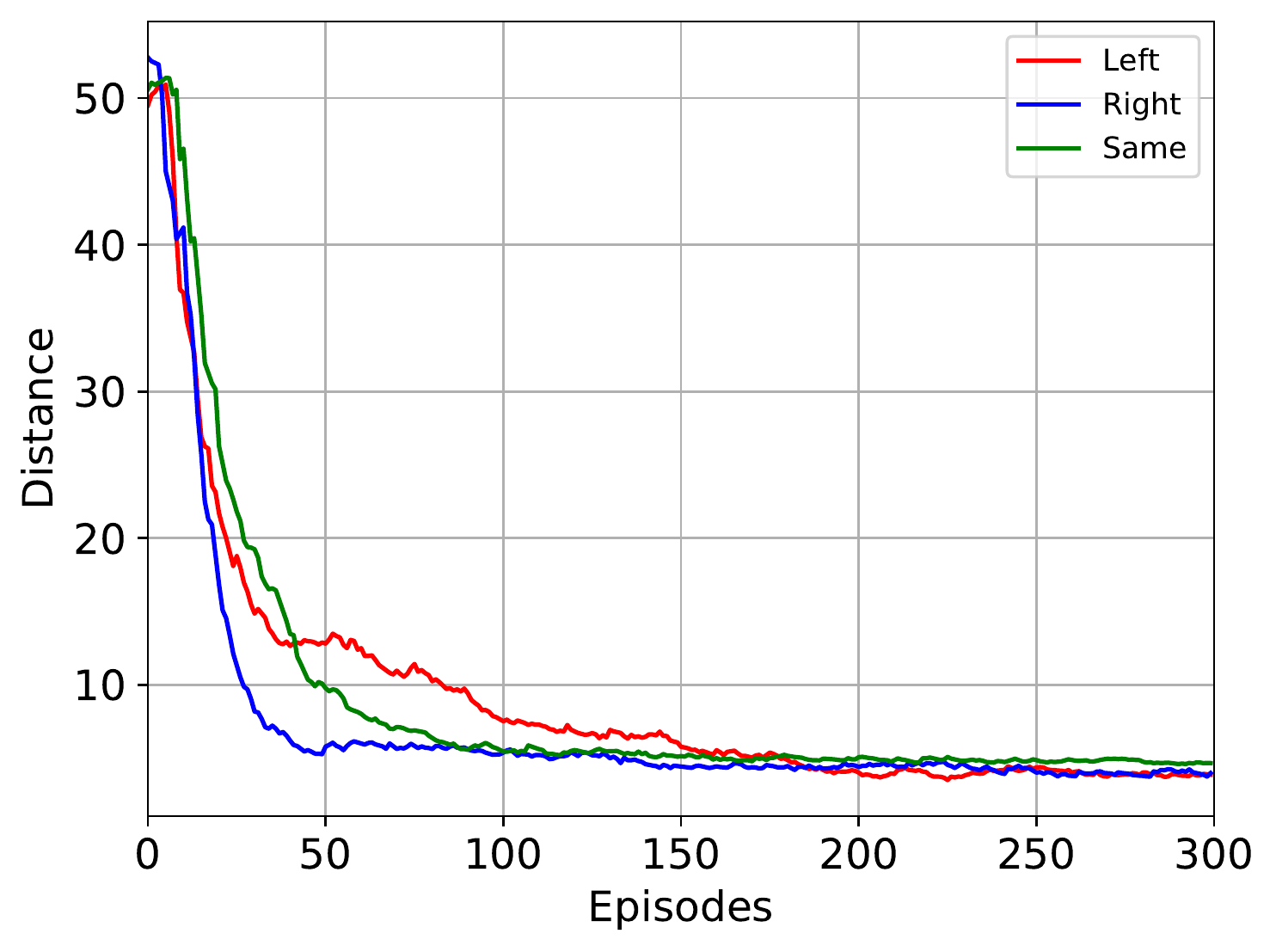} \label{fig:EstimatedDistanceDeterministic}}
\subfloat[Noisy signal from (\ref{fig:ProbabilityMemoryDeterministic}).]{\includegraphics[width=0.33\textwidth]{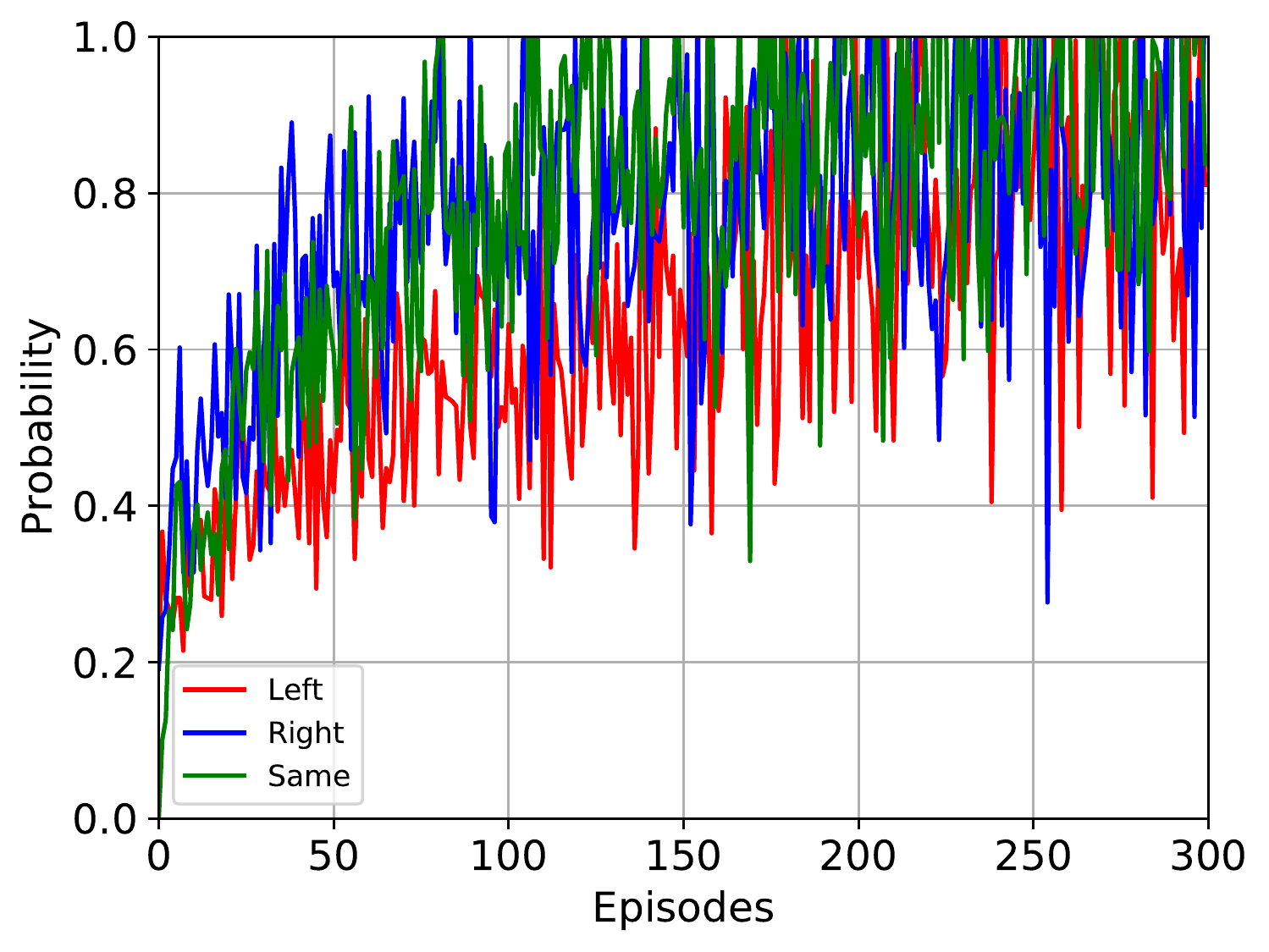} \label{fig:ProbabilityNoisedDeterministic}}
\\
\subfloat[Memory-based approach.]{\includegraphics[width=0.33\textwidth]{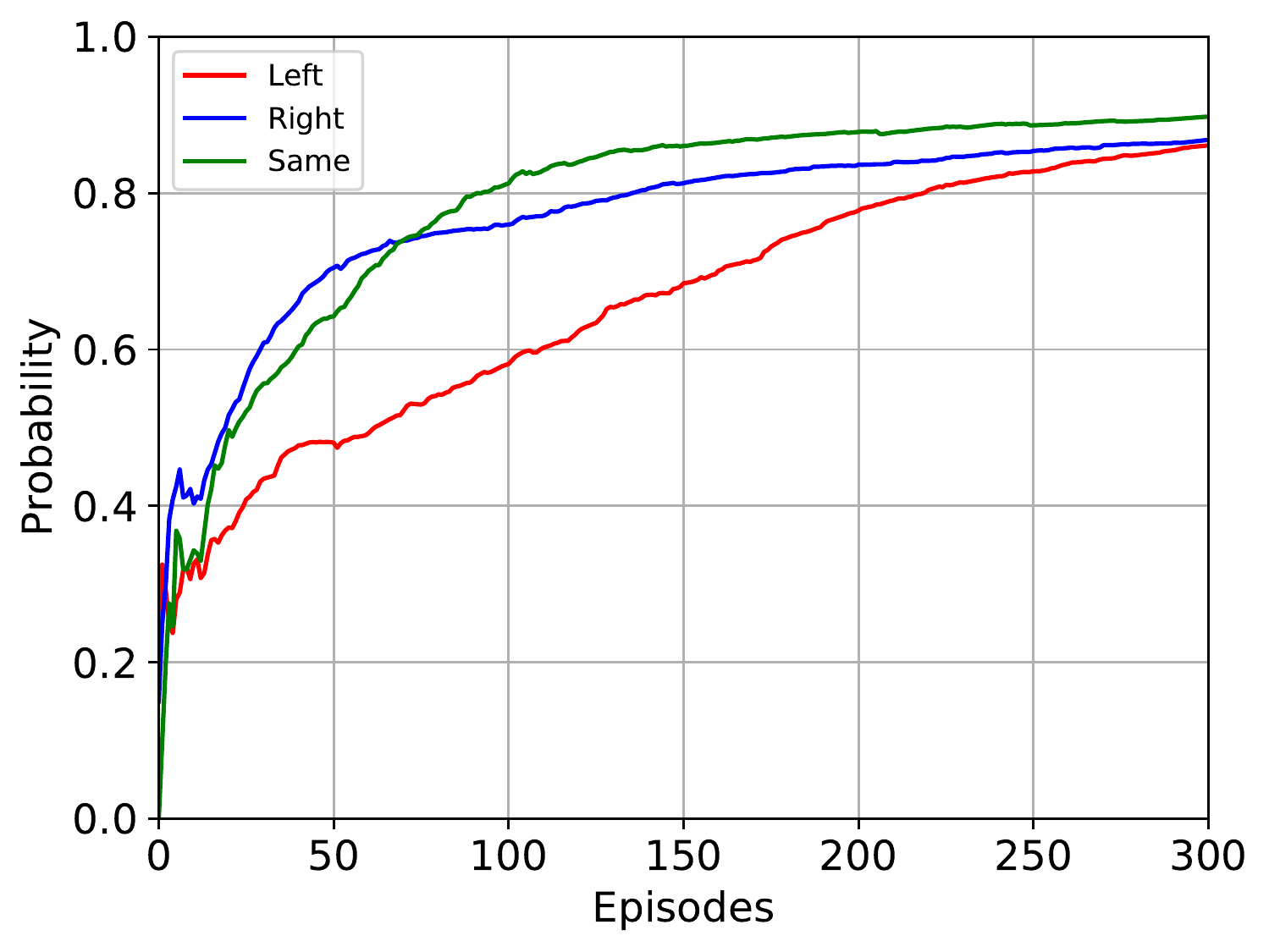} \label{fig:ProbabilityMemoryDeterministic}}
\subfloat[Learning-based approach.]{\includegraphics[width=0.33\textwidth]{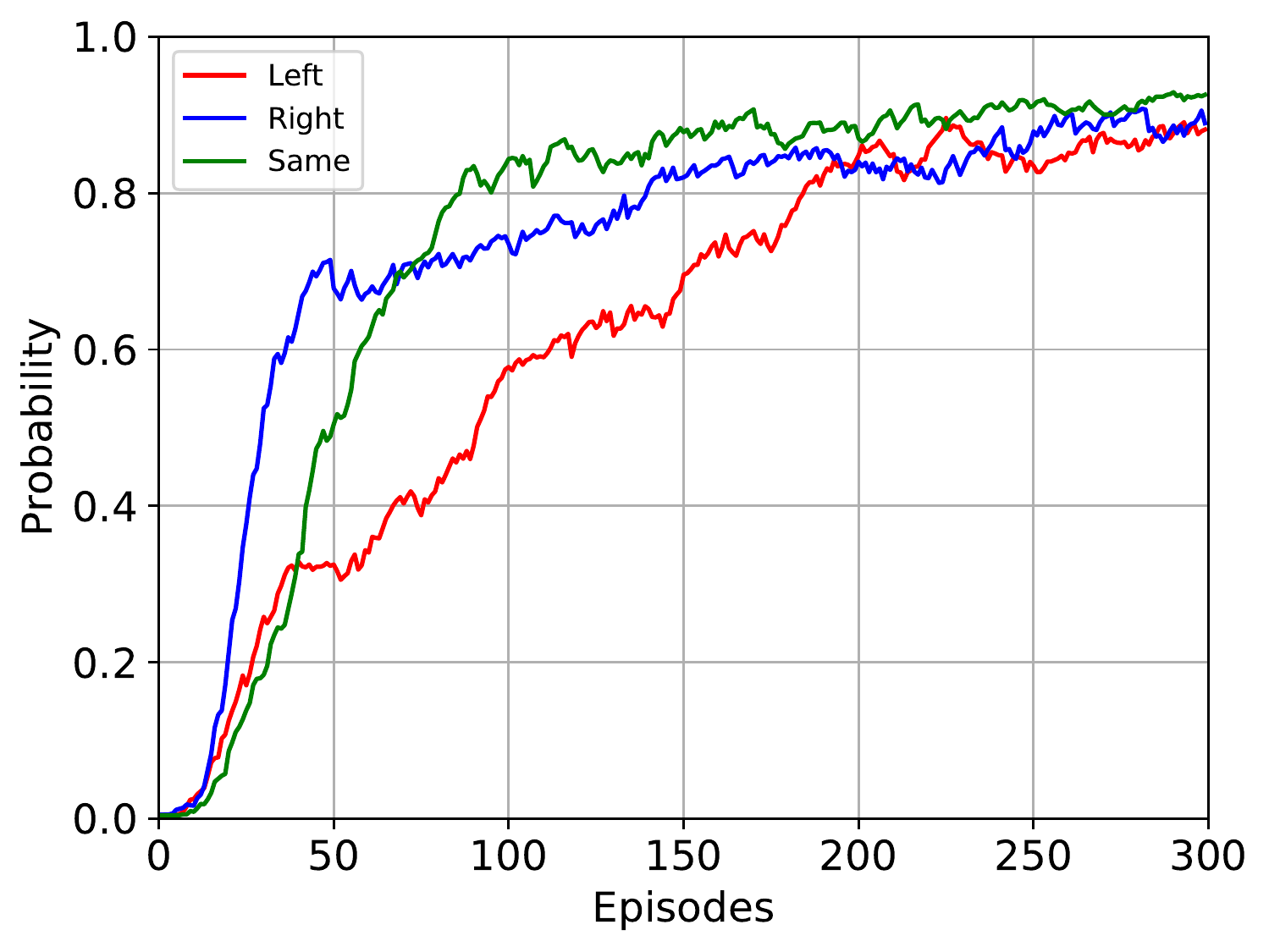} \label{fig:ProbabilityLearningDeterministic}}
\subfloat[Introspection-based approach.]{\includegraphics[width=0.33\textwidth]{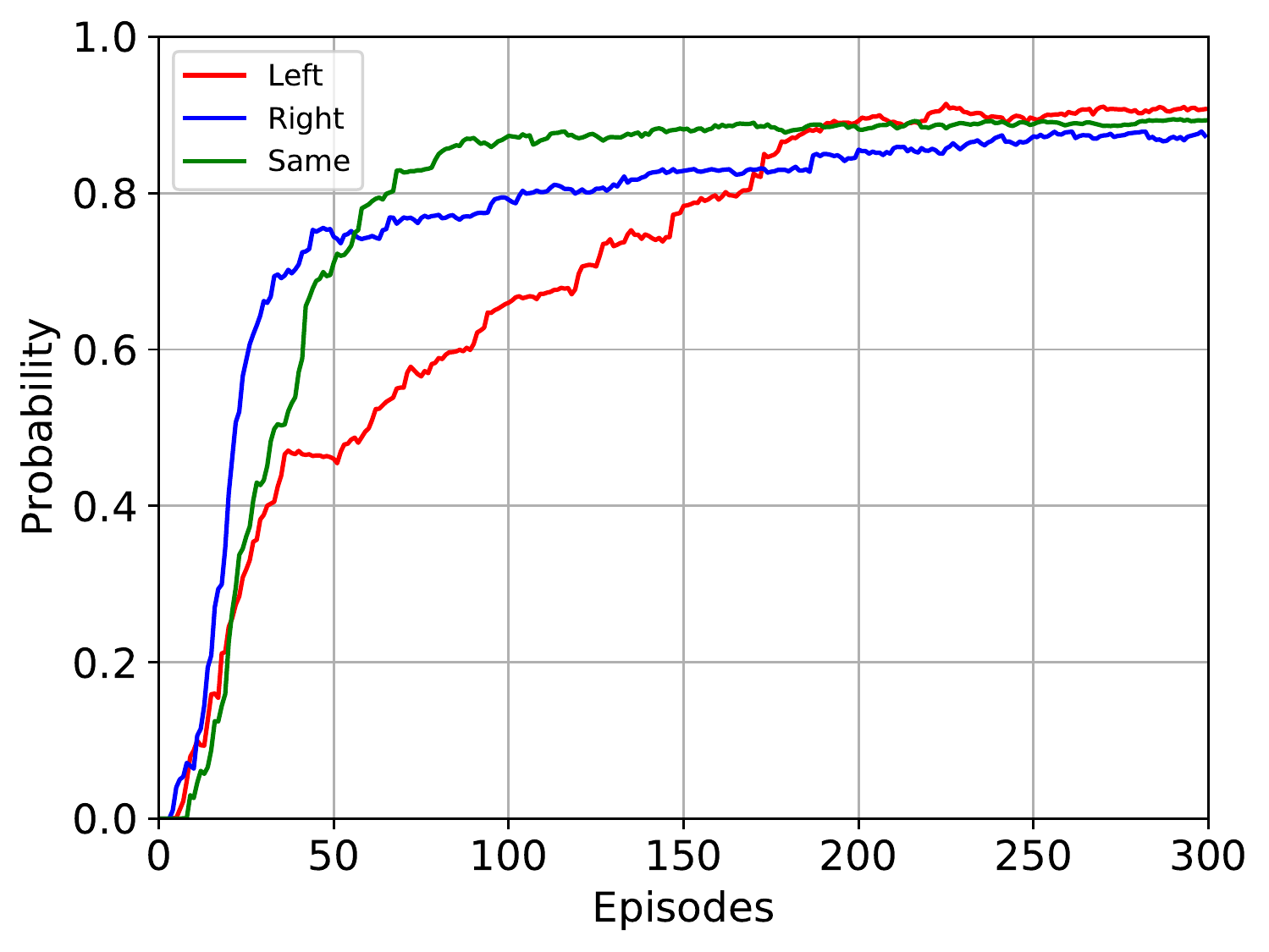} \label{fig:ProbabilityIntrospectionDeterministic}}
\\
\subfloat{\includegraphics[width=0.35\textwidth]{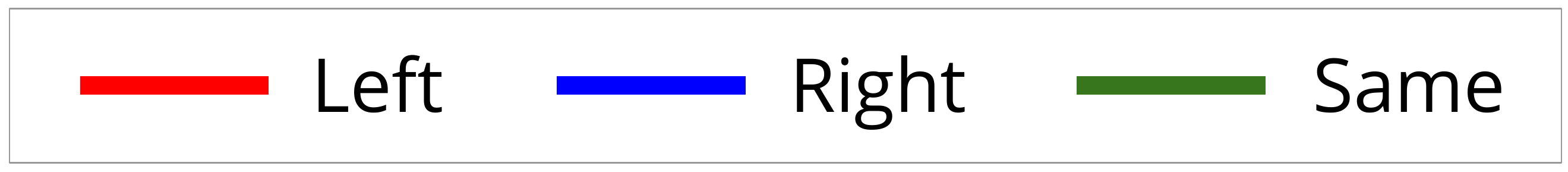}}
\caption{Deterministic robot navigation task. Results are shown from the initial state $s_0$ for all possible actions. 
  During the first episodes the agent prefers going to the right room as appreciated in the Q-values, however, all the probabilities converge to similar values after 300 episodes. 
  Over the episodes, the estimated distance $n$ gets closer to the minimum number of actions.}
\label{fig:DeterministicRobotNavigation}
\end{figure*}

Fig.~\ref{fig:QValuesDeterministic} shows the Q-values obtained over $300$ episodes for the actions of moving to the left $a_L$, to the right $a_R$, and staying in the same room $a_S$  from the initial state $s_0$.
It is possible to observe that during the first episodes the agent prefers to perform $a_R$ as a consequence of the collected experience which is shown by the blue line.
However, as the learning improves, the three actions converge to similar Q-values above $0.6$.

In Fig.~\ref{fig:EstimatedDistanceDeterministic} can be seen the estimated distance $n$ (according to Eq. \eqref{Eq:n}) from the initial state $s_0$ to the reward by performing the three available actions.
Contrary to the Q-values, the distance needed to reach the reward decreases over time, starting with more than $50$ actions and reaching values close to the minimum.
It can be seen that action right $a_R$ converges faster since the estimated distance is computed according to the agent's experience using the Q-values and the reward.
Since the distance $n$ is obtained from the Q-values, it can be produced for any of the proposed approaches, however, we use it specifically to compute the probability of success with the introspection-based approach.

In Fig.~\ref{fig:ProbabilityMemoryDeterministic} are shown the probabilities of success for the memory-based approach taking the different possible actions from the initial state $s_0$, i.e., the probability of successfully finishing the task choosing any path from room $0$.
The three possible actions from this state, i.e., go to the left room $a_L$, go to the right room $a_R$, and stay at the same room $a_S$ are shown using red, blue, and green respectively.
In the first episodes, any possible action has a very low probability of success since the agent still does not know how to navigate appropriately and, therefore, often selects an action that leads it out of the floor.
Over the episodes, the agent tends to follow the path to its right to reach the goal state, however, after 300 episodes all the probabilities converge to a similar value as the agent collects enough knowledge in all paths.

By using the probability of success as a source of information for a non-expert end-user, the RL agent can more easily explain its behavior in terms of its goals and why at a certain point of the learning process one action may be preferred instead of others.
For instance, after 50 learning episodes, the computed probabilities of success are 48.09\%, 70.46\%, and 64.24\% for actions $a_L$, $a_R$, and $a_S$ respectively.
Therefore, the robot might justify its behavior as follow: \textit{`From the initial room, I chose to move to the right because it had the biggest probability of successfully finishing the task'}. 
Since the computed probabilities of success using the memory-based approach are obtained directly from the performed actions during the episodes, following we use these results to compare to both: the learning-based approach and the introspection-based approach.
Moreover, we use a noisy signal obtained from the memory-based approach as a  control group.
For all the approaches, including the noisy signal, we compute the Pearson's correlation to measure the similarity between the approaches as well as the mean square error (MSE).

Fig.~\ref{fig:ProbabilityLearningDeterministic} shows~ the~ probability~ of~ success~ for~ the three possible actions from the initial state $s_0$ using the learning-based approach.
Similarly to the memory-based approach, the probabilities show the agent initially prefers to perform the action for moving to the right $a_R$, however, as the previous case, after the learning process, all the probabilities converge to similar values close to 90\% of success.

The probabilities of success using the introspection-based approach are shown in Fig.~\ref{fig:ProbabilityIntrospectionDeterministic}.
As before, the possible actions are shown from the initial state $s_0$.
The evolution of the probabilities behaves similarly over the episodes as in previous approaches.
Initially, the agent favors the action to the right room but the three actions reach a similar probability of success after training.

Equivalently as shown by the learning-based approach, the introspection-based approach in the first episodes gets a probability of success equal to zero $P_s=0$.
This initial behavior is due to the fact that the probability of success $P_s$ using the learning-based approach is computed in a similar way as the Q-values, which is updating the $\mathbb{P}$-values inside the $\mathbb{P}$-table.
Likewise, the estimated probability of success $\hat P_s$ using the introspection-based approach is computed from the Q-values as it is a numerical transformation from the estimated distance $n$.

Overall, the three proposed approaches have similar behavior when using deterministic transitions reaching similar results in terms of the final probabilities of success from the initial state $s_0$ and the evolution over the learning process.
To further analyze the similarity between the proposed approaches we compute the Pearson's correlation as well as the MSE with respect to the memory-based approach.
Additionally, we have used a control group of probabilities of success as a noisy signal from the memory-based approach using 20\% of white noise $(M=1, SD=0.2)$.
We have used that amount of noise since we want to create control data from our baseline approach which are different enough from the original probabilities and, at the same time, distinguishable from each other. 
However, in this work, we do not test how tolerant our approaches are to respond to possible noise.
The resultant noisy probabilities of success can be seen in Fig.~\ref{fig:ProbabilityNoisedDeterministic}.

\begin{figure}[!hbt]
  \centering
  \includegraphics[width=1\linewidth]{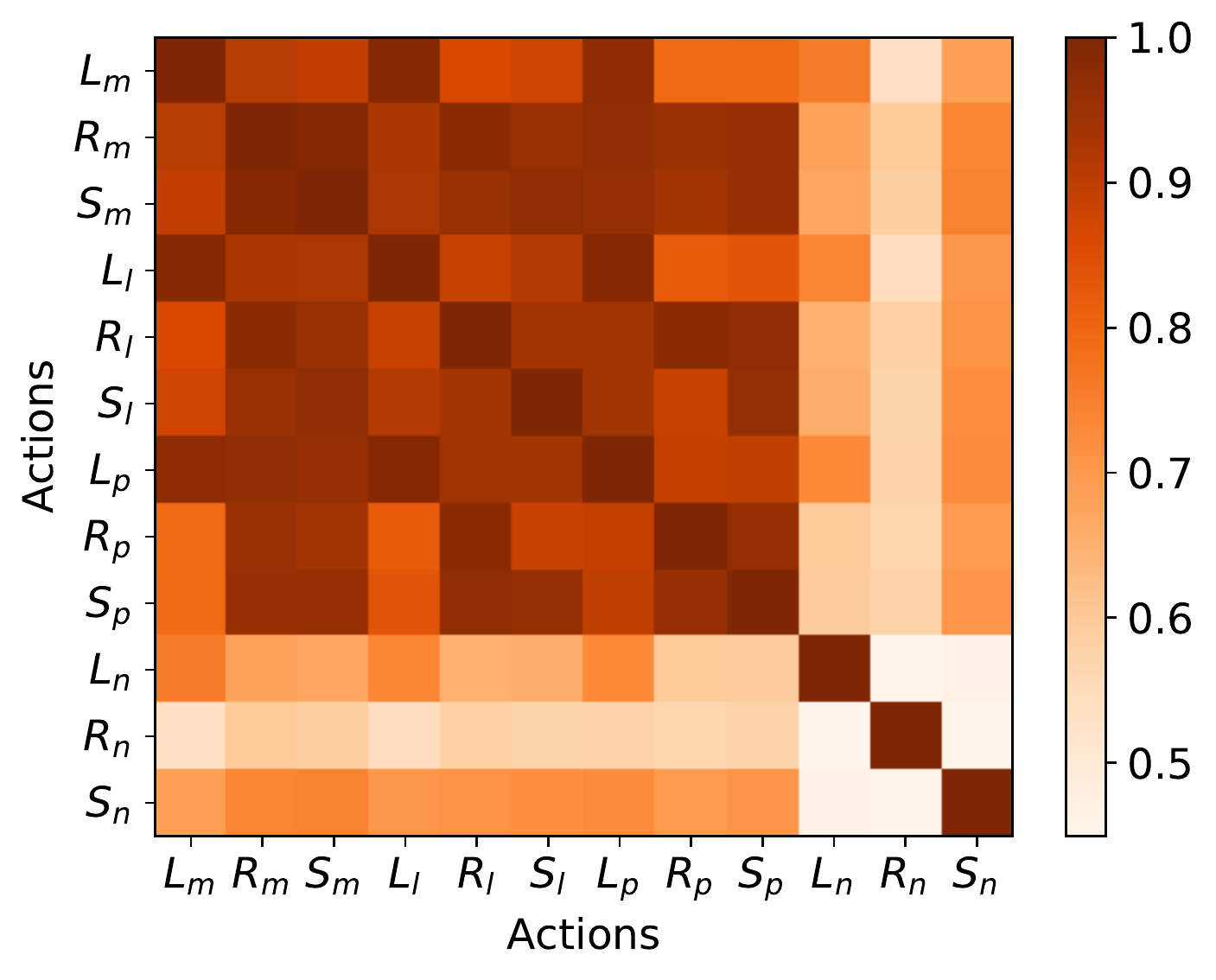}
  \caption{Pearson's correlation between the probabilities of success for all the approaches considering the three possible actions from the initial state $s_0$ in the deterministic robot navigation task.
  All the approaches obtain a similar behavior with the exception of the noisy signal obtained from the memory-based approach and used as a control group.}
  \label{fig:CorrelationDeterministic}
\end{figure}

In Fig.~\ref{fig:CorrelationDeterministic} is shown the correlation matrix for all the approaches.
In the figure, axes are the different actions from the initial state $s_0$ for each proposed method.
The uppercase letter refers to the action and the lowercase one refers to the method.
Thus, $L$, $R$, and $S$ are for the actions of going to the left, to the right, and staying in the same room, whereas $m$, $l$, $p$, and $n$ are for memory-based, learning-based, introspection-based, and noisy approaches respectively.
The figure shows that there is a high correlation between the three proposed approaches, while in our noisy control group the values of the correlations are much lower in comparison.

Moreover, Table~~\ref{tab:MSEDeterministic} shows the MSE between the memory-based approach and all the other approaches.
It can be seen that the introspection-based approach approach has the least amount of errors in relation to the memory-based benchmark, obtaining an MSE lower than $0.01$ for all possible actions, which is achieved with much lower memory-usage than the memory-based approach.

\begin{table}
  \centering
  \caption{MSE for all the proposed approaches using deterministic transitions against the memory-based approach. All shown actions are performed from the initial state $s_0$.}
  \label{tab:MSEDeterministic}
  \begin{tabular}{lccc}
    \hline
    \textbf{Deterministic app.} & $\boldsymbol{a_L}$ & \textbf{$\boldsymbol{a_R}$} & \textbf{$\boldsymbol{a_S}$} \\
    \hline
    Learning-based & 0.0117 & 0.0110 & 0.0178 \\
    Introspection-based & 0.0095 & 0.0065 & 0.0083 \\
    Noisy Memory-based & 0.0187 & 0.0245 & 0.0226 \\
    \hline
  \end{tabular}
\end{table}

\subsection{Stochastic robot navigation task}

In this section, we have performed the same robot navigation task but using stochastic transitions instead as is the case in many real-world scenarios. 
In our scenario, to use stochastic transitions means that the RL agent may perform the action $a$ from the state $s_i$ to the state $s_j$ and may reach the intended state $s_j$ with a transition probability $p_t < 1$, or more precisely $p_t= 1-\sigma$, with $\sigma \in [0,1]$ (if $\sigma=0$ deterministic transitions are used, i.e., no stochasticity) taking into consideration the defined transition function.
We have introduced a transition probability $p_t=0.9$, or $\sigma=0.1$.
In other words, a $10\%$ of stochasticity in order to test how coherent are the possible explanations extracted from all the proposed approaches.

\begin{figure*}[!hbt]
\centering
\subfloat[Q-values.]{\includegraphics[width=0.33\textwidth]{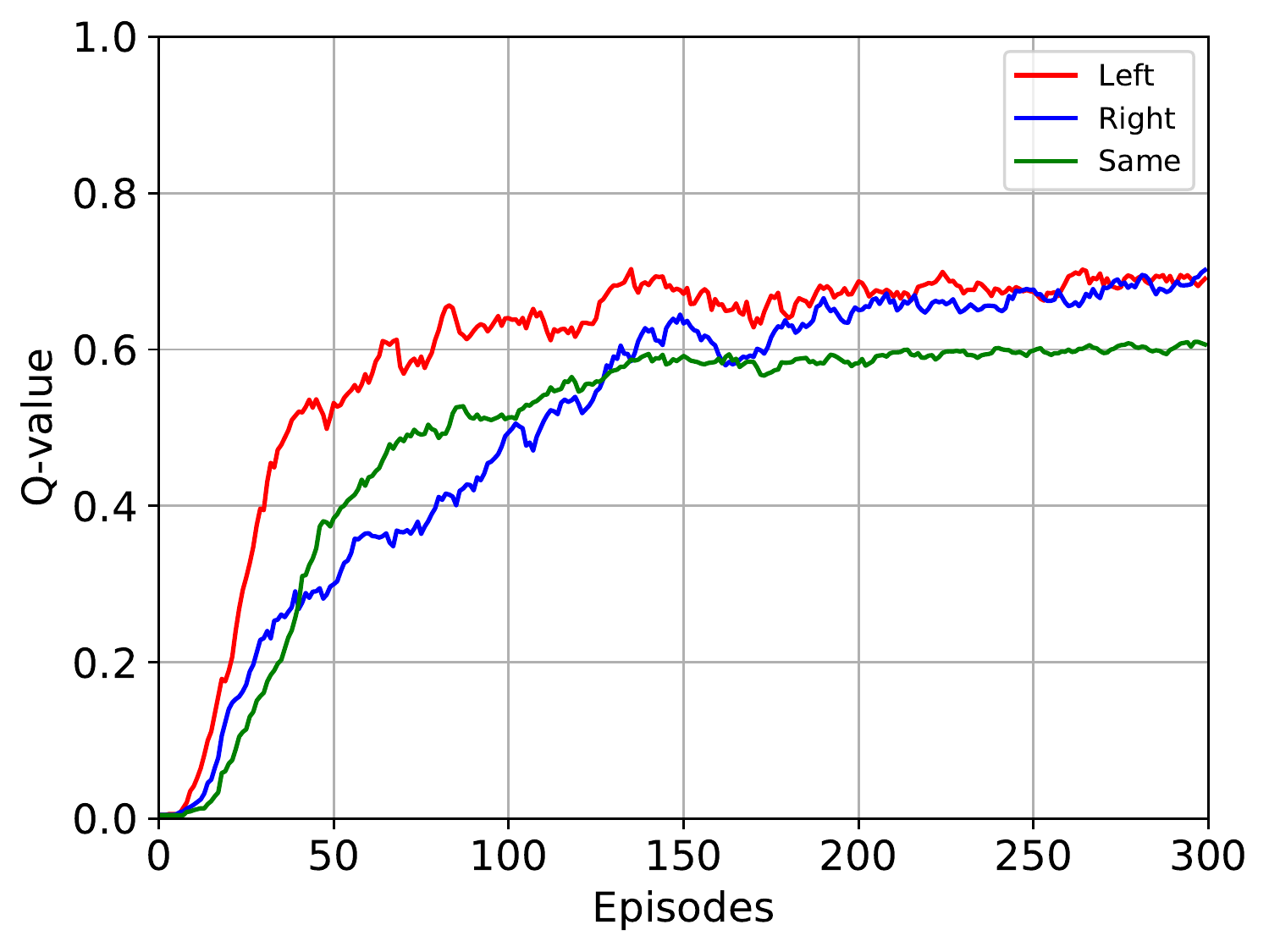} \label{fig:QValuesStochastic}} 
\subfloat[Estimated distance n.]{\includegraphics[width=0.33\textwidth]{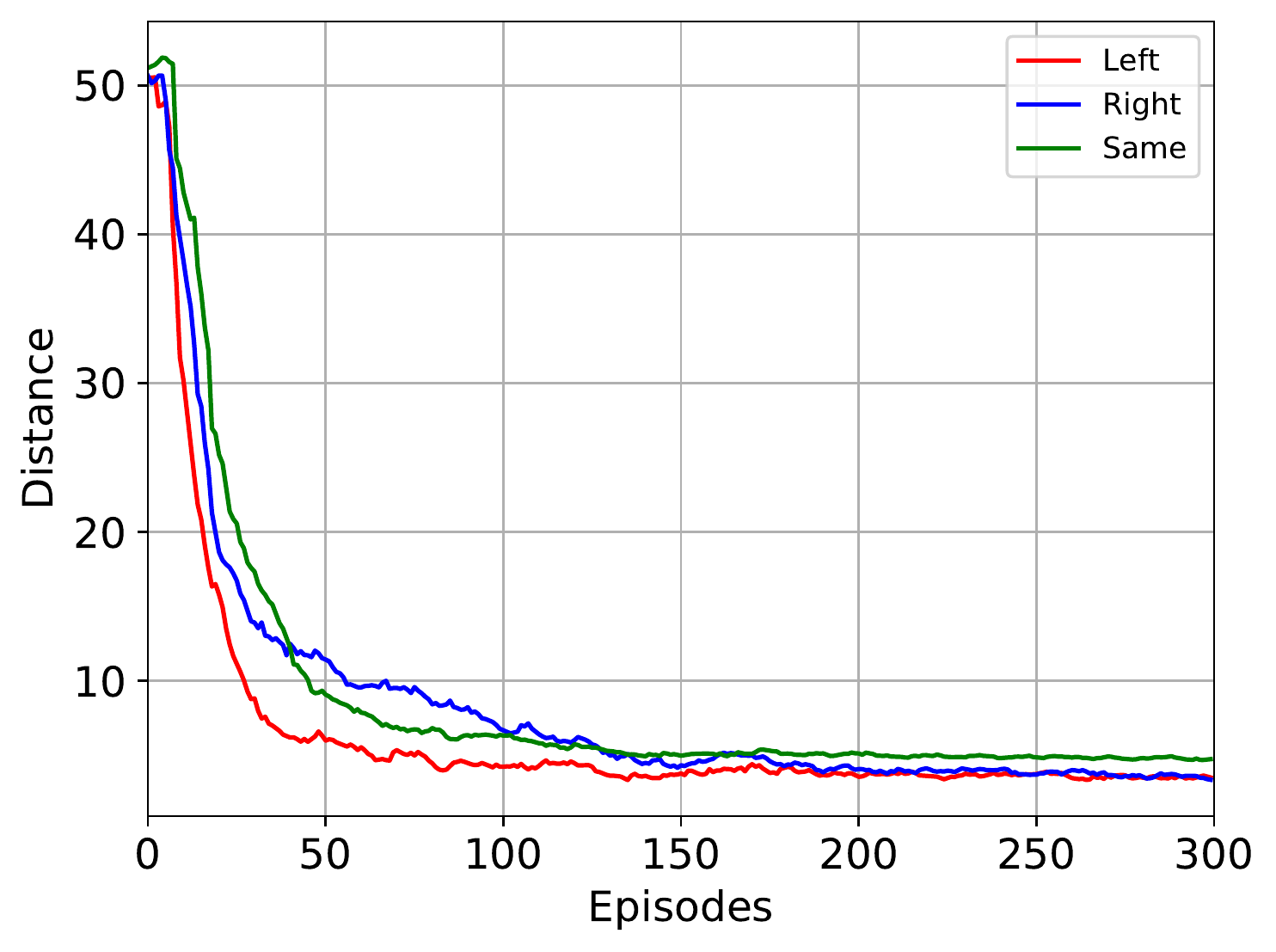} \label{fig:EstimatedDistanceStochastic}}
\subfloat[Noisy signal from (\ref{fig:ProbabilityMemoryStochastic}).]{\includegraphics[width=0.33\textwidth]{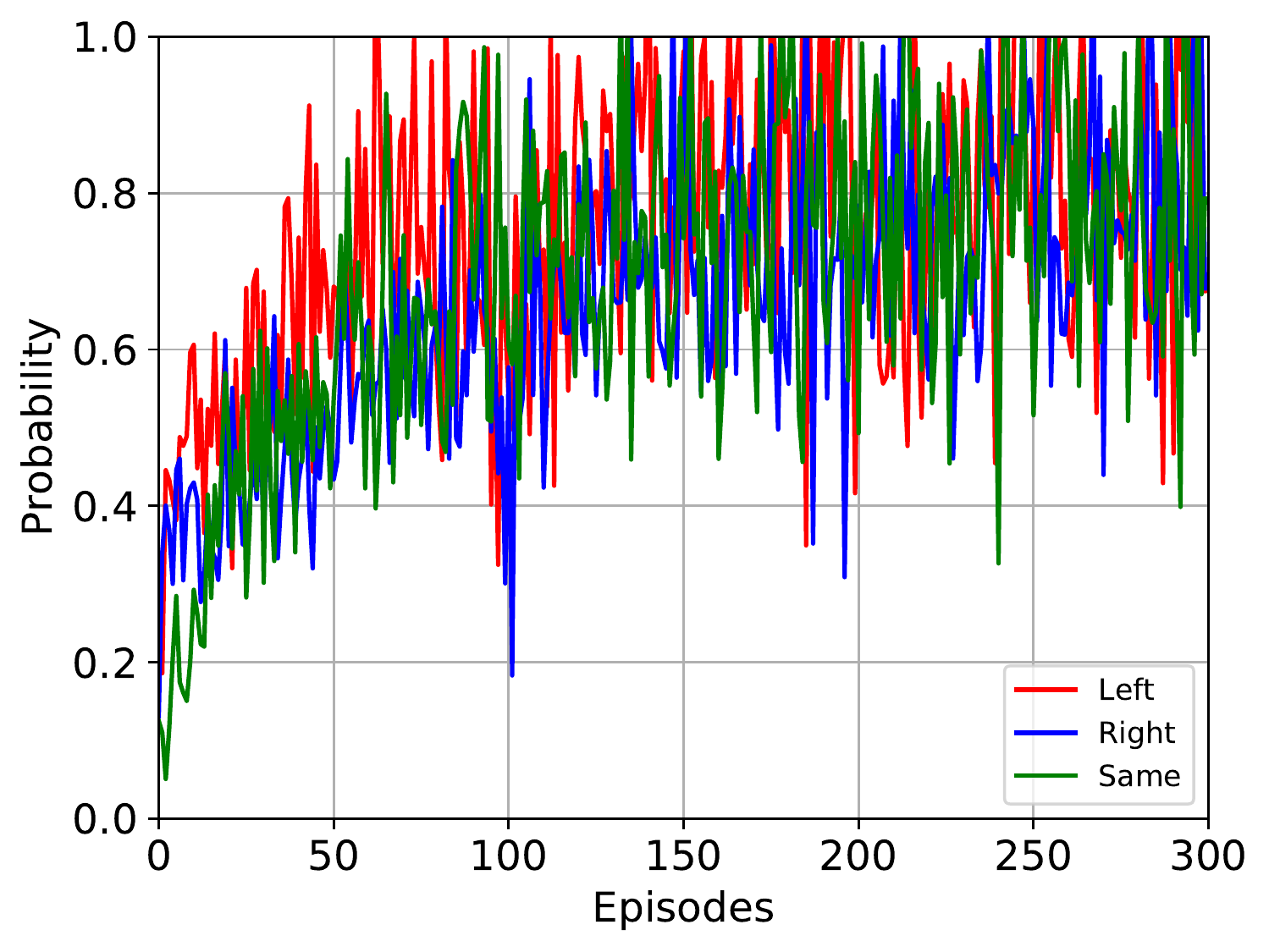} \label{fig:ProbabilityNoisedStochastic}}
\\
\subfloat[Memory-based approach.]{\includegraphics[width=0.33\textwidth]{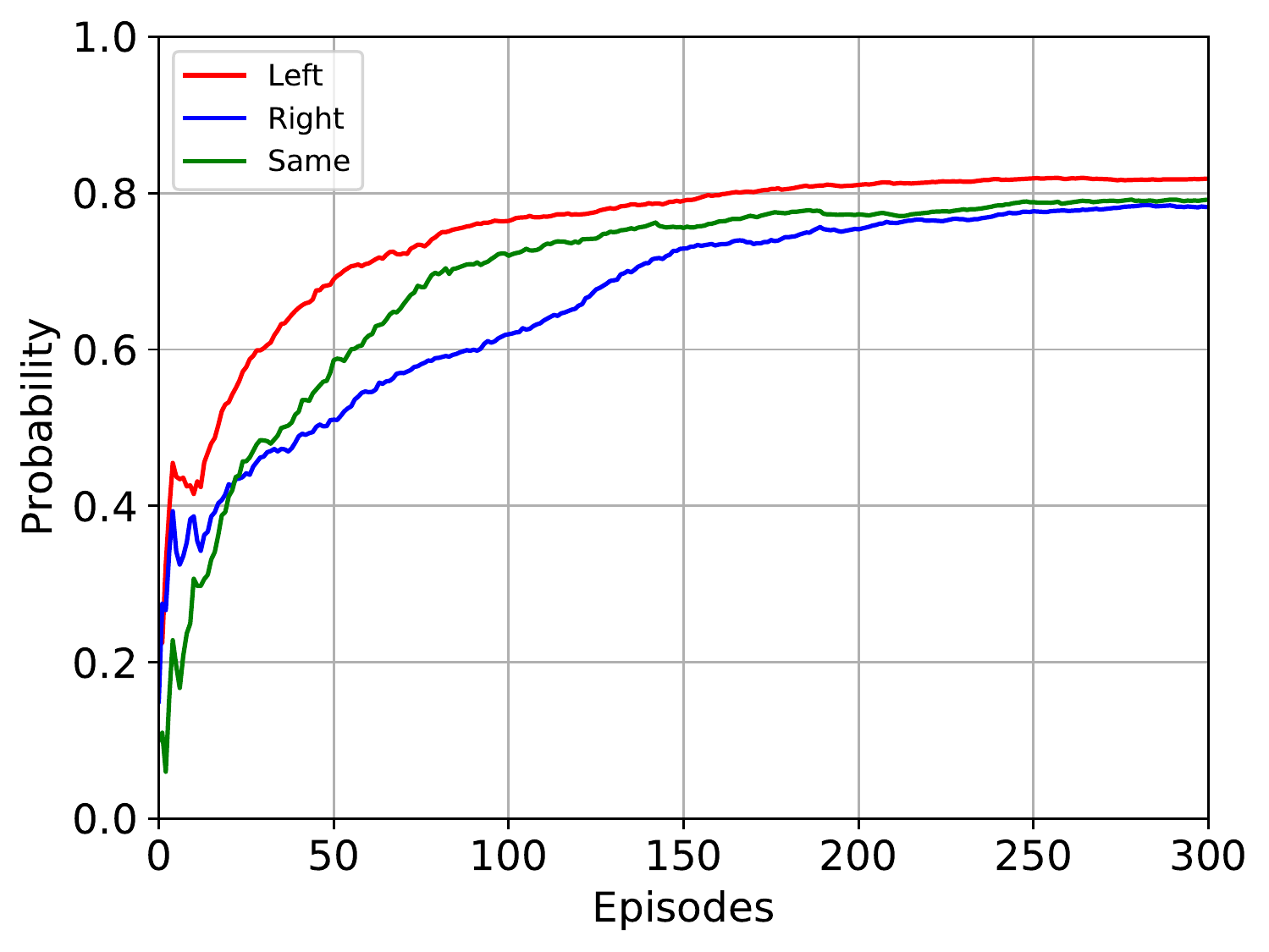} \label{fig:ProbabilityMemoryStochastic}}
\subfloat[Learning-based approach.]{\includegraphics[width=0.33\textwidth]{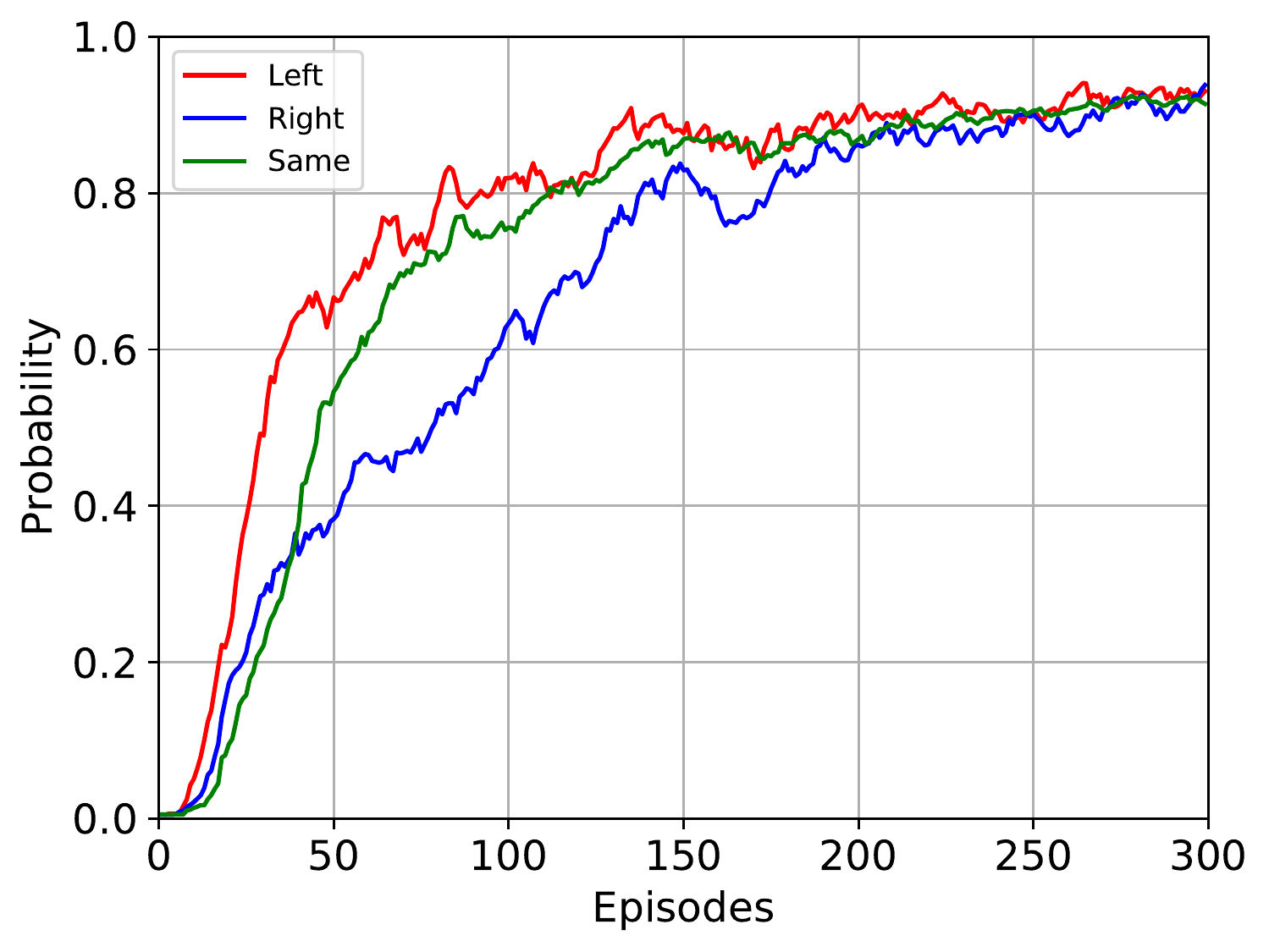} \label{fig:ProbabilityLearningStochastic}}
\subfloat[Introspection-based approach.]{\includegraphics[width=0.33\textwidth]{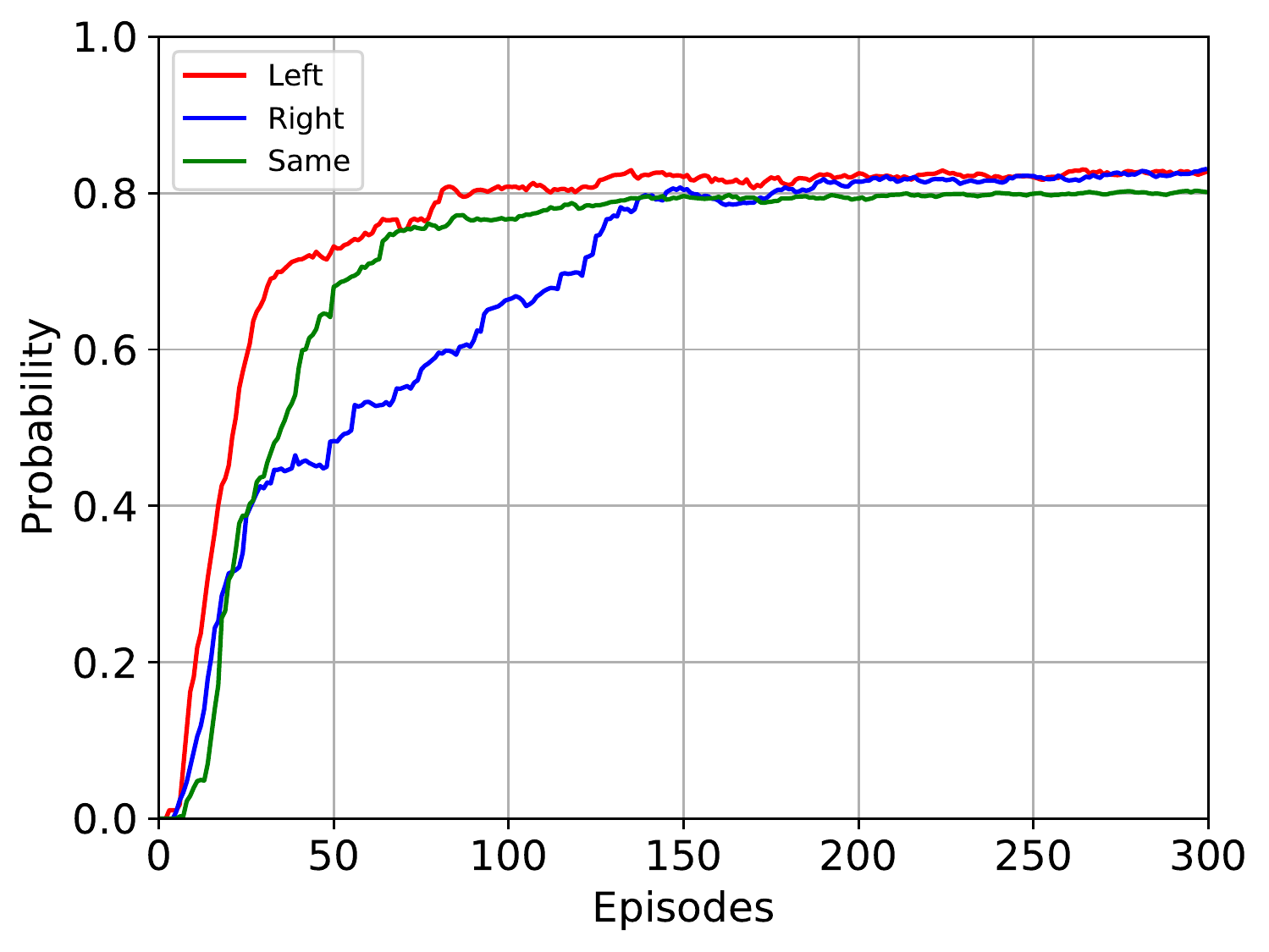} \label{fig:ProbabilityIntrospectionStochastic}}
\\
\subfloat{\includegraphics[width=0.35\textwidth]{figures/LabelXRL.pdf}}
\caption{Stochastic robot navigation task. Results are shown from the initial state $s_0$ for all possible actions. 
  In this case, the agent favors going to the left room at the beginning of the training, however, as in the previous case, for each approach all the probabilities converge to similar values after 300 episodes.
  The estimated probabilities of success are lower than the deterministic case, which is due to the introduced $\sigma$-value stochasticity.
  In the learning-based approach, due to the fact of using a discount factor equals to one ($\gamma=1$), the agent becomes more foresighted and take into consideration all the possible future reward, which leads to slightly higher probabilities of success.
  The estimated distance $n$ converges close to the minimum for all actions after the learning process.
  }
\label{fig:StochasticRobotNavigation}
\end{figure*}

Fig.~\ref{fig:QValuesStochastic} shows the obtained Q-values after the learning process.
The actions shown also correspond to the three possibilities, i.e., $a_L$, $a_R$, and $a_S$, from the initial state $s_0$.
Similarly to the use of deterministic transitions, the Q-values converge to similar values after 300 episodes, however, in this case the agent favors the action of going left in the first episodes.
Certainly, this is not due to the use of stochastic transitions, but rather the agent in this experiment explored initially that path, which can lead to diverse exploration experiences over different learning processes.

In Fig.~\ref{fig:EstimatedDistanceStochastic} is shown the estimated distance $n$ in terms of actions to the reward.
By using stochastic transitions the distances also decrease over time getting close to the minimal amount of actions.
In this case, the action of going to the right $a_R$ needs more time to converge since this path is explored later as shown in the Q-values.

Fig.~\ref{fig:ProbabilityMemoryStochastic} shows the probabilities of success during the learning process from the initial state $s_0$ using stochastic transitions and the memory-based approach.
In this case, the agent initially exhibits more experience taking the path to the left; however, after 300 episodes, similarly as before, the probabilities converge to a similar value.
Although using stochastic transitions lead to a less overall probability of success, in comparison to the deterministic robot navigation task, the agent is still able to explain in these terms the reasons for its behavior during the learning episodes.
In this case, after 50 learning episodes, the computed probabilities of success are 68.99\%, 51.02\%, and 58.65\% for actions $a_L$, $a_R$, and $a_S$ respectively.
Therefore, according to the experience collected by the robot, the behavior might be justified as: \textit{`From the initial room, I chose to go to the left because it had the biggest probability of success'}. 
Like the previous case, we have used the memory-based approach as a baseline to compare the other proposed approaches, since these probabilities are obtained directly from the actual robot experience collected in the episodic memory.

The probabilities of success for the three possible actions from the initial state $s_0$ using the learning-based approach are shown in Fig.~\ref{fig:ProbabilityLearningStochastic}.
Like using the memory-based approach, at the beginning of the training, the agent shows more experience by following the path to the left, but also the three actions converge after training.
In this case, the probabilities of success converge to a slightly higher amount in comparison to the memory-based approach.
This is due to the fact that these probabilities are computed using the $\mathbb{P}$-values from the $\mathbb{P}$-table.
These values are updated according to Eq. \eqref{Eq:learning}, where we set the discount factor $\gamma=1$ and, therefore, the agent is more foresighted taking into account all possible future rewards.

The estimated probabilities of success using the introspection-based approach are shown in Fig.~\ref{fig:ProbabilityIntrospectionStochastic}.
The probabilities are for the three possible actions from the initial state $s_0$ as well.
The introspection-based approach has a similar behavior as the memory-based approach, converging to similar values after the learning process when using stochastic transitions.
The estimated probabilities are computed from the Q-values using the Eq.~\ref{Eq:PsBound} and, therefore, it can be similarly seen that the agent preferred the path to the left at the beginning of the training.

\begin{figure}[!hbt]
  \centering
  \includegraphics[width=1\linewidth]{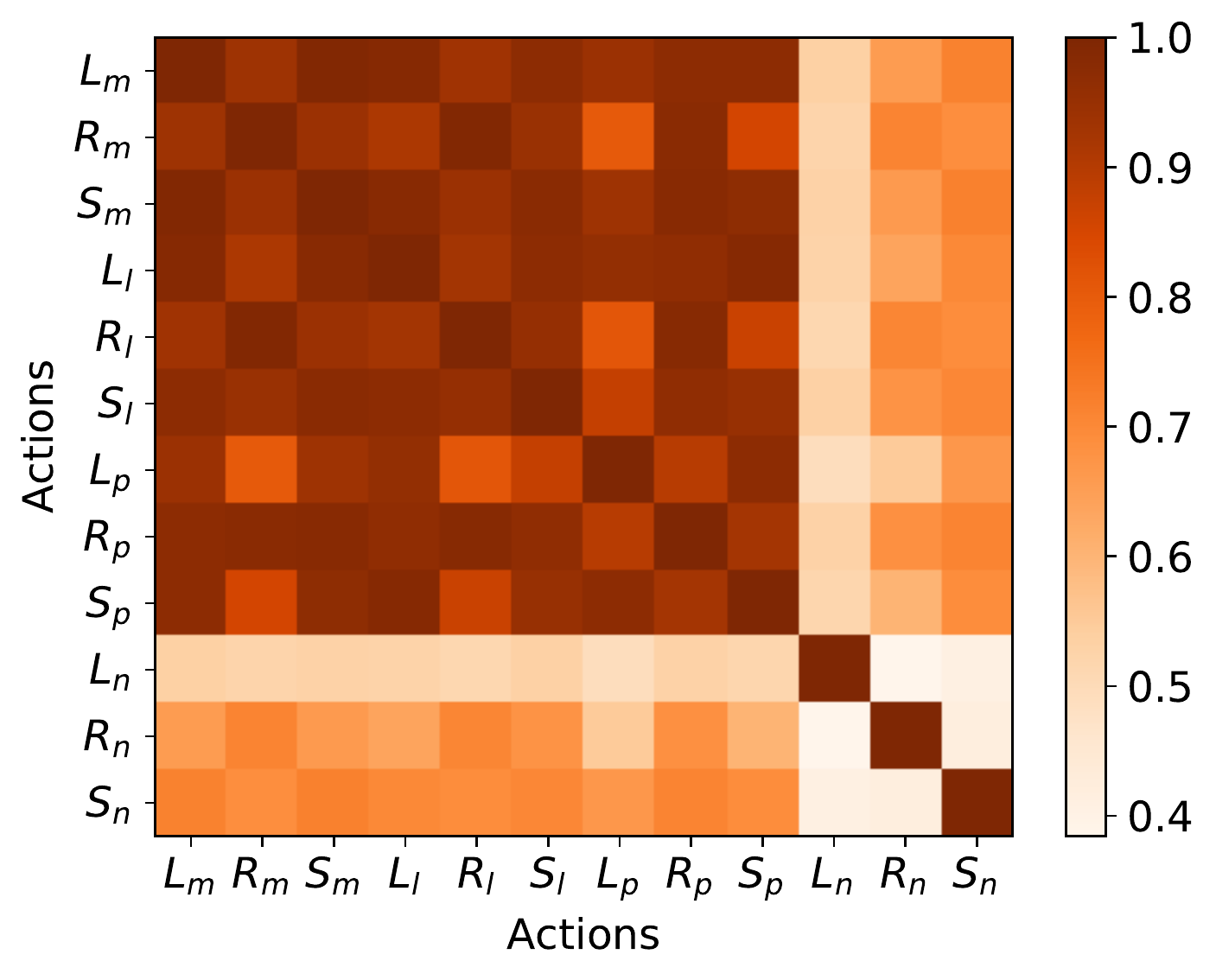}
  \caption{Pearson's correlation between the probabilities of success for all the approaches considering the three possible actions from the initial state $s_0$ in the stochastic robot navigation task.
  Although the correlation is lower in comparison to the deterministic approach due to the stochastic transitions, all the approaches obtain a similar behavior with the exception of the noisy signal.}
  \label{fig:CorrelationStochastic}
\end{figure}

As in the deterministic case, we have also used a noisy signal in the stochastic robot navigation task as a control group.
We obtained the noisy signal from the memory-based approach since this is computed from the actual agent's interaction during the learning process.
Once again, we have added a 20\% of white noise using a normal distribution with media $1$ and standard deviation $0.2$, i.e., $(M=1, SD=0.2)$.
The noisy signal can be seen in Fig.~\ref{fig:ProbabilityNoisedStochastic}.

As in the deterministic robot navigation task, we also compute the Pearson's correlation as well as the MSE to analyze the similarity between the obtained probabilities of success $P_s$.
In Fig.~\ref{fig:CorrelationStochastic} is shown the correlation matrix for all the approaches.
The axes contain the different possible actions from the initial state $s_0$ for each proposed method.
As mentioned, the uppercase letter refers to the action and the lowercase letter refers to the method.
Although the correlations are lower in comparison to the deterministic robot navigation task due to the use of stochastic transitions, the figure still shows that there is a high correlation between the three proposed approaches, in opposite to the noisy control group, where the values of the correlations are lower in comparison.

Furthermore, Table~\ref{tab:MSEStochastic} shows the MSE between the memory-based approach and the other approaches using stochastic transitions.
It is observed that once again the introspection-based approach is the most similar to the memory-based approach, obtaining errors lower than $0.007$ for all possible actions, which is also achieved using much less memory in comparison to the memory-based approach.

\begin{table}
  \centering
  \caption{MSE for all the proposed approaches using stochastic transitions against the memory-based approach. All shown actions are performed from the initial state $s_0$.}
  \label{tab:MSEStochastic}
  \begin{tabular}{lccc}
    \hline
    \textbf{Stochastic app.} & $\boldsymbol{a_L}$ & \textbf{$\boldsymbol{a_R}$} & \textbf{$\boldsymbol{a_S}$} \\
    \hline
    Learning-based & 0.0153 & 0.0171 & 0.0161\\
    Introspection-based & 0.0058 & 0.0068 & 0.0045\\
    Noisy Memory-based & 0.0271 & 0.0166 & 0.0218\\
    \hline
  \end{tabular}
\end{table}

\begin{figure*}[!hbt]
\centering
\subfloat[Q-values.]{\includegraphics[width=0.5\textwidth]{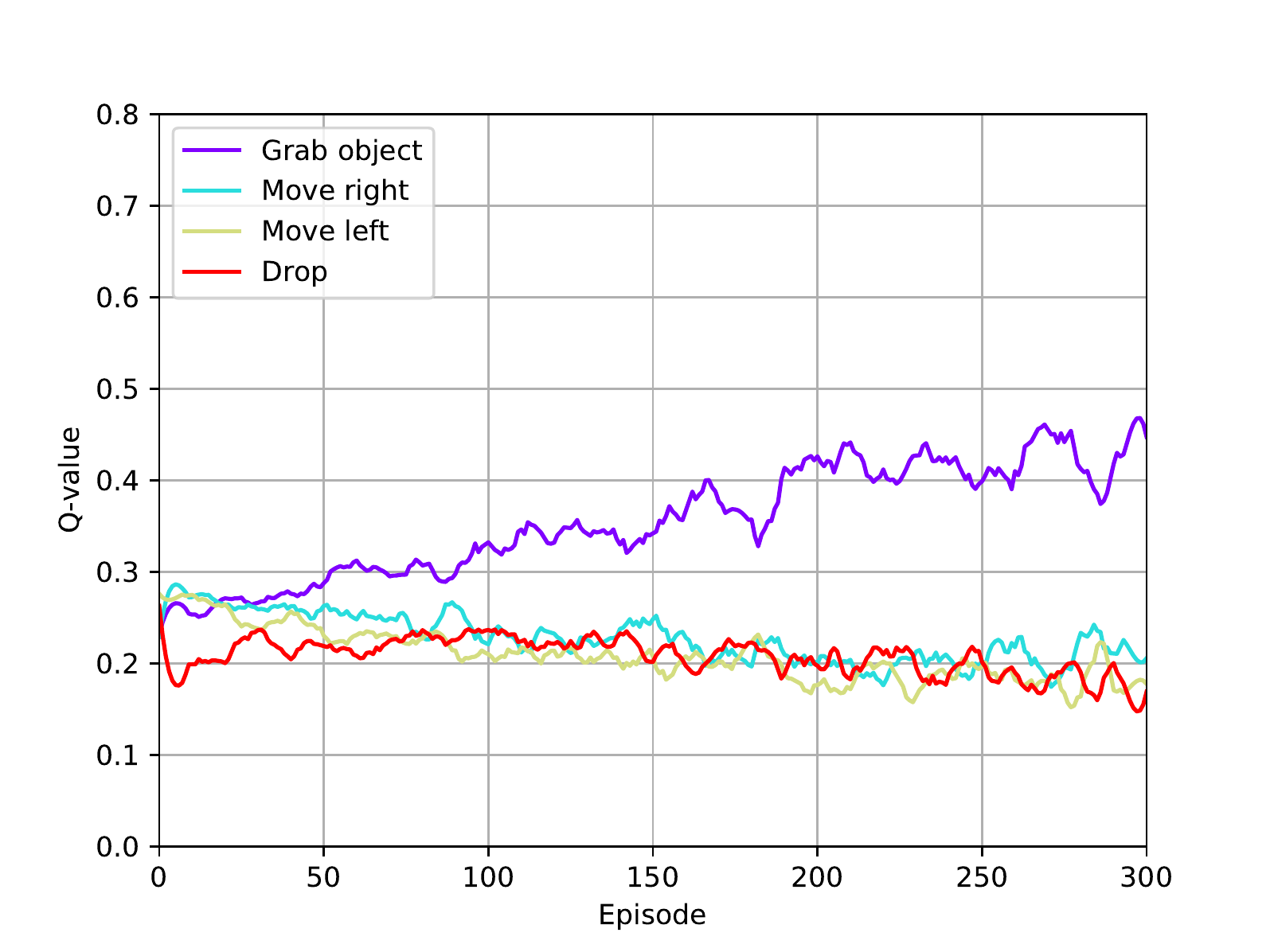} \label{fig:SortingQValues}} 
\subfloat[Introspection-based approach.]{\includegraphics[width=0.5\textwidth]{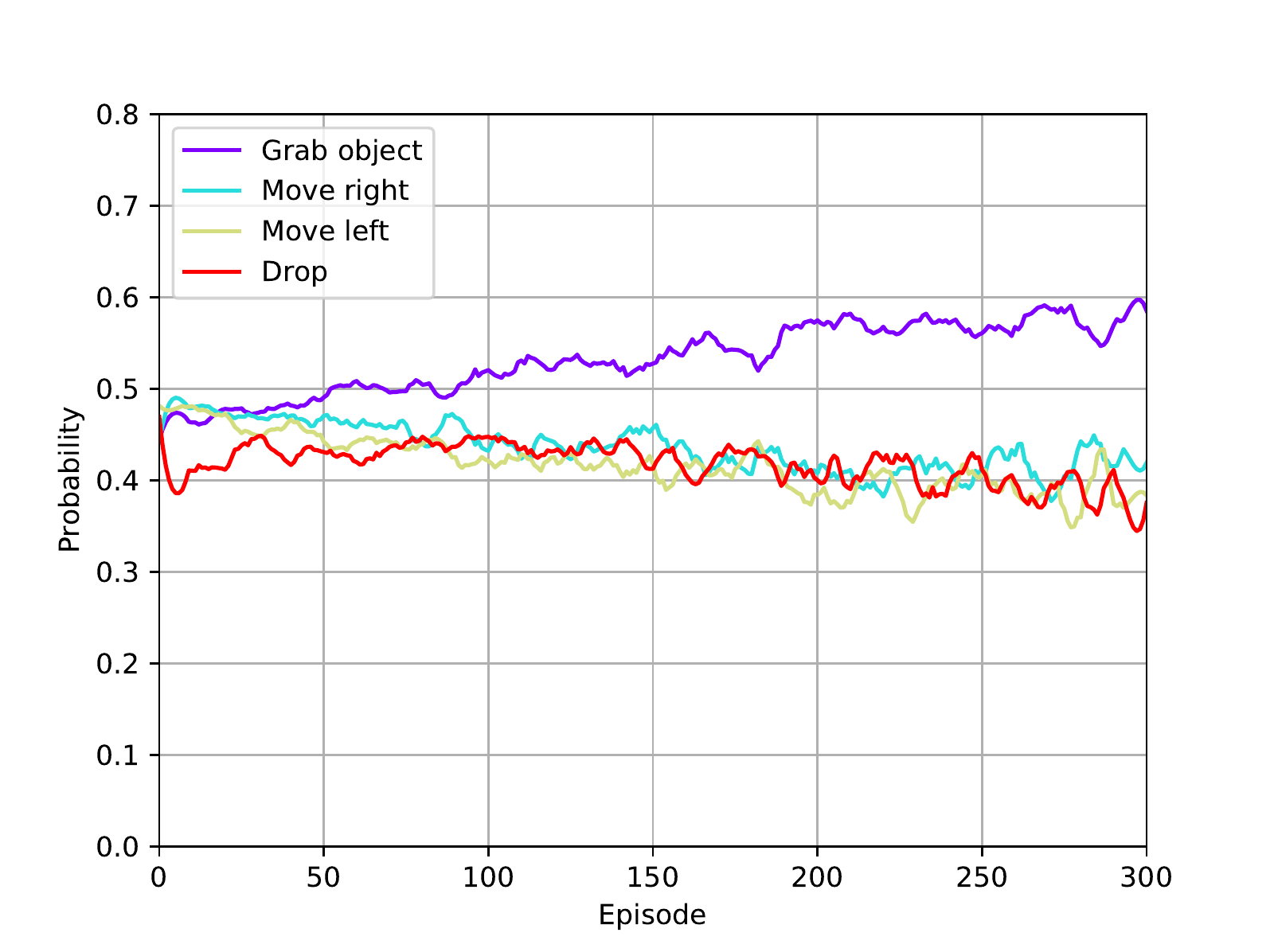} \label{fig:SortingProbabilities}}
\\
\subfloat{\includegraphics[width=0.6\textwidth]{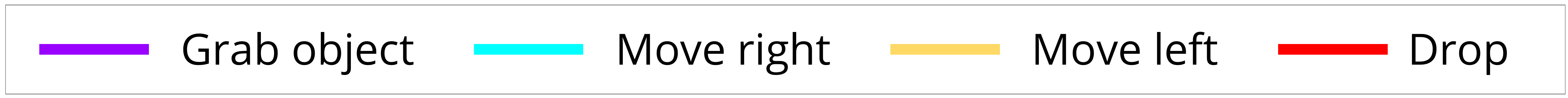}}
\caption{Continuous sorting object task. Results are shown from the initial state for all possible actions. 
  In this case, the agent favors the action of grabbing an object in order to complete the task faster. 
  The estimated probabilities by the introspection-based approach are computed with the logarithmic transformation directly from the Q-values.
  Due to the transformation, the probabilities may allow producing better explanations to non-expert end-users.
  }
\label{fig:SortingObjectResults}
\end{figure*}

\subsection{Continuous sorting object task}

In this section, we show the results obtained in a continuous task. 
The continuous sorting object task has been used along with the introspection-based approach.
As demonstrated in the previous section, all the proposed approaches are equivalent in terms of the probabilities of success obtained, however, not all of them are suitable for all problem representations.
Moreover, as aforementioned, the introspection-based approach is more efficient ($~O(1)$) in terms of memory use and time required for learning.  
In this problem, using the memory-based approach is not feasible since no tabular representation is possible; moreover, the learning-based approach would require the training of an additional neural network as a function approximator.

Figure~\ref{fig:SortingQValues} shows the Q-values from the initial state for the 4 possible actions averaged for 10 agents.
The results are smoothed using a Savitzky–Golay filter~\cite{savitzky1964smoothing} with a window of length 15 and a polynomial of order 3 to fit the data.
As mentioned, the Q-values are in the reward function domain and the dominant initial action during the 300 learning episodes is 'grab object', which indeed leads to finish the task faster.
The Q-values for this action vary from 0.25 to 0.45 during the training, thus, if they are directly used to generate an explanation, it would be meaningless for non-expert users.

The probabilities of success are computed directly from the Q-values using the logarithmic transformation introduced in the introspection-based approach. 
Figure~\ref{fig:SortingProbabilities} shows the probabilities of success from the initial state for the 4 possible actions.
The results are averaged for 10 agents and smoothed using the Savitzky–Golay filter once again with a window of 15 and the data fitted with a polynomial of order 3.
In this case, it is observed that estimated probabilities for the action 'grab object' vary from values close to 0.5 to 0.6.
This respectively means $50\%$ and $60\%$ of probability for successfully completing the task when performing this action from the initial state.
In this regard, an explanation elaborated in such a way may be much more intuitive to follow by non-experts interacting with the robot.

Similarly, counterfactual arguments may be given.
For instance, after 300 learning episodes, a user may ask in the initial state 'why the action move right or move left have not been chosen by the agent'.
The robot may intuitively answer this question using the computed probability of success for the actions.
Thus, an answer may be \textit{'I have selected the action grab object because doing so, I have $59\%$ chances of sorting all the objects successfully, while moving left I have only $38\%$ probability of being successful'}.


\subsection{Real-world scenario}

As previously mentioned in Section \ref{sec:RealScenario}, we have performed a simple real-world test involving 
two different situations.
%
In the first situation, the robot performed the action $a_R$ (move to the right) from the initial position (room 0).
We showed to the user one of the following explanations:
(i)~\textit{I moved to the right because it has a Q-value of  0.744}, or (ii)~\textit{I moved to the right because it has a 95.8\% probability of reaching the table}.
%
%
%
In the second situation, the robot performed the action $a_L$ (move to the left) that led the robot to the goal position (room 5).
We showed to the user one of these explanations:
(i)~\textit{I moved to the left because it has a Q-value of 1}, or (ii)~\textit{I moved to the left because it has a 100\% probability of reaching the table}.

Both Q-values and probabilities of success $P_s$ have been computed with the introspection-based approach.
As non-expert end-users have no understanding of the meaning of a Q-value necessarily, we hypothesize that an explanation using the probability of success should bring more clear ideas in order to understand the robot's behavior.
However, the situations are different.
While in the first situation the robot is still far from the reward, in the second situation the robot is initially just one action away from the goal position.
Therefore, we consider that end-users may have a general better understanding of the second situation since this needs only one action to complete the task. 
Therefore in this situation, explanations may make more sense to users regardless if they are shown using Q-values or the probability of success.
Another aspect that may affect the explainability judgment is the previous experience in the current scenario.
Whatever situation is experienced first by an end-user, this may bias the next situation, since they have already seen the robot acting and may have a better perception of the robot's behavior. 
Nevertheless, we have performed this test as a proof of concept and we do believe that further human-robot interactive experiments are needed in order to draw meaningful conclusions.
For instance, it is important to test the approaches including end-users with different backgrounds, aiming at people with no experience in machine learning.
Additionally, including people with different levels of educational background (e.g., English language, knowledge in math, etc.) would also benefit future tests since our explanations are intended for non-expert end-users as widely as possible.

\section{Conclusions}

In this work, we have presented an explainable robotic system to supply understandable human-like explanations to an end-user in a human-robot environment. 
The experimental scenarios have been carried out using a robot simulator where 2 robotic tasks have been implemented.
Additionally, a real-world proof of concept has been carried out in order to validate the probability of success as a valid metric for explainability purposes. 
The proposed approaches do not focus on speeding up the learning process nor on generating automatic explanations, but rather on looking for a plausible and practical means of explaining the robot's behavior during the decision-making process. 
For this purpose, we have used goal-driven explanations instead of state-based explanations.

The proposed approaches estimate the probability of success for each action, which in turn allows the agent to explain the robot's decision to non-expert end-users.
By describing decisions in terms of the probability of success, the end-user will have a clearer idea about the robot's decision in each situation using human-like language.
On the contrary, using Q-values to explain the behavior, the end-users will not necessarily obtain a straightforward comprehension unless they have prior knowledge about reinforcement learning or machine learning techniques. 

We have implemented three approaches with different characteristics.
First, the memory-based approach uses an episodic memory to save the interaction with the environment from which it computes the probability of success.
Second, the learning-based approach utilizes a $\mathbb{P}$-table where the values of the probability of success are updated as the agent collects more experience during the learning process.
Third, the introspection-based approach computes the estimated probability of success directly from the Q-values by performing a numerical transformation.
The proposed approaches differ in both the amount of memory needed and the kind of RL problem representation where they could be used. 


Using either deterministic or stochastic transitions in the robot navigation task, the obtained results show that the proposed approaches accomplish similar behavior and converge to similar values, which is also verified through the high correlation level and the MSE computed. 
Overall, from the similarity exposed by the three proposed approaches, the learning-based approach and the introspection-based approach represent a plausible choice for replacing the memory-based approach to compute the probability of success, using fewer memory resources and being an alternative to other RL problem representations. 
In this regard, the introspection-based approach is the one requiring fewer resources in comparison to the 2 others.
Additionally, we have demonstrated that the introspection-based approach is easily scaled up to continuous scenarios by implementing the visual-based sorting object task.

Although we have used simulated scenarios, an additional important aspect to consider in explainable robotic systems is real-world scenarios in which human-robot trust may be critical, such as medical or safety scenarios, or when decisions are taken based on biased data~\cite{setchi2020explainable}.
If a robot is required to give an explanation during the first learning stages, the RL agent may not have collected all the necessary information to answer properly.
In other words, the robot may provide an incorrect or poor explanation that although it may be the best to its current knowledge, still it is not necessarily correct after the optimal policy has been learned.
If this explanation seems coherent to the end-user, this may create an unjustified trust feeling and lead to inappropriate decisions, and in critical cases exposing the human to unsafe or risky situations.
If the explanation is obviously wrong, this may deteriorate the trust relationship and the human counterpart may not tolerate that kind of answer in the future~\cite{sakai2021explainable}.
A comprehensive explanation is sometimes impossible given that the underlying algorithms may be too complex to explain to non-expert end-users.
Sometimes, it is not possible due to privacy breaches.
Nevertheless, it is important to focus on explaining the robot's behavior to keep it accountable and working according to the laws~\cite{dawson2019artificial}.


As future works, we see reward decomposition~\cite{juozapaitis2019explainable} as a good alternative to include additional reward signals, such as more complex penalties, or to address multi-objective problems.
In this regard, each meaningful part of the reward signal may be decomposed in order to compute the different independent probabilities of success and generate explanations for each of them. 
It is also important to test how much impact different values of stochastic transitions, represented by the $\sigma$ parameter, especially considering that in real-world scenarios stochasticity may vary between states.

We also plan to use a real-world robot scenario in order to automatically generate explanations to be given to non-expert end-users.
Using a real-world scenario, we plan to perform a user study to measure the effectiveness of the probability of success as a metric to enhance the trust in robotic systems.

\begin{figure*}[!hbt]
\centering
\subfloat[Q-values.]{\includegraphics[width=0.33\textwidth]{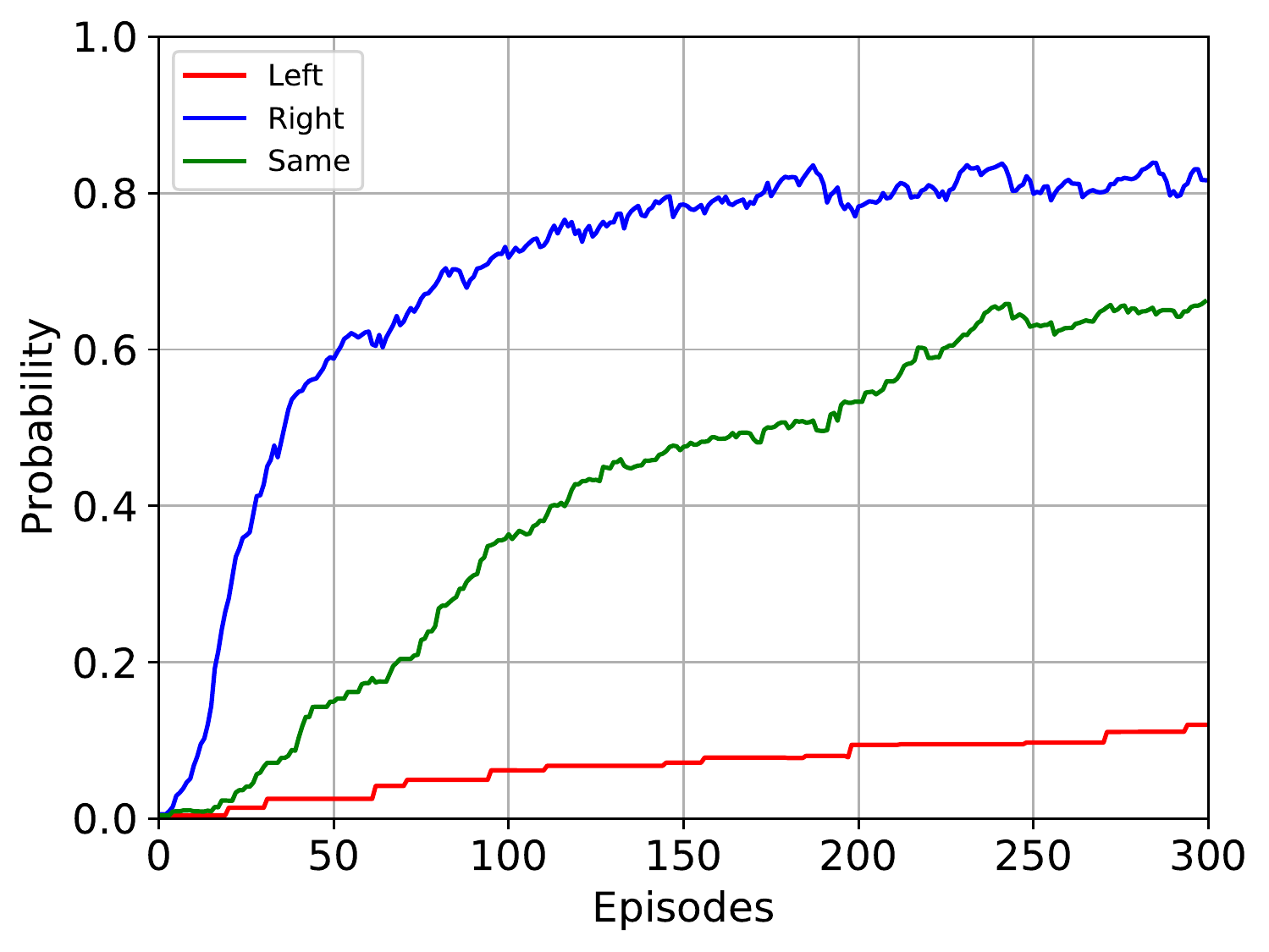} \label{fig:QValuesStochasticS1}} 
\subfloat[Memory-based approach.]{\includegraphics[width=0.33\textwidth]{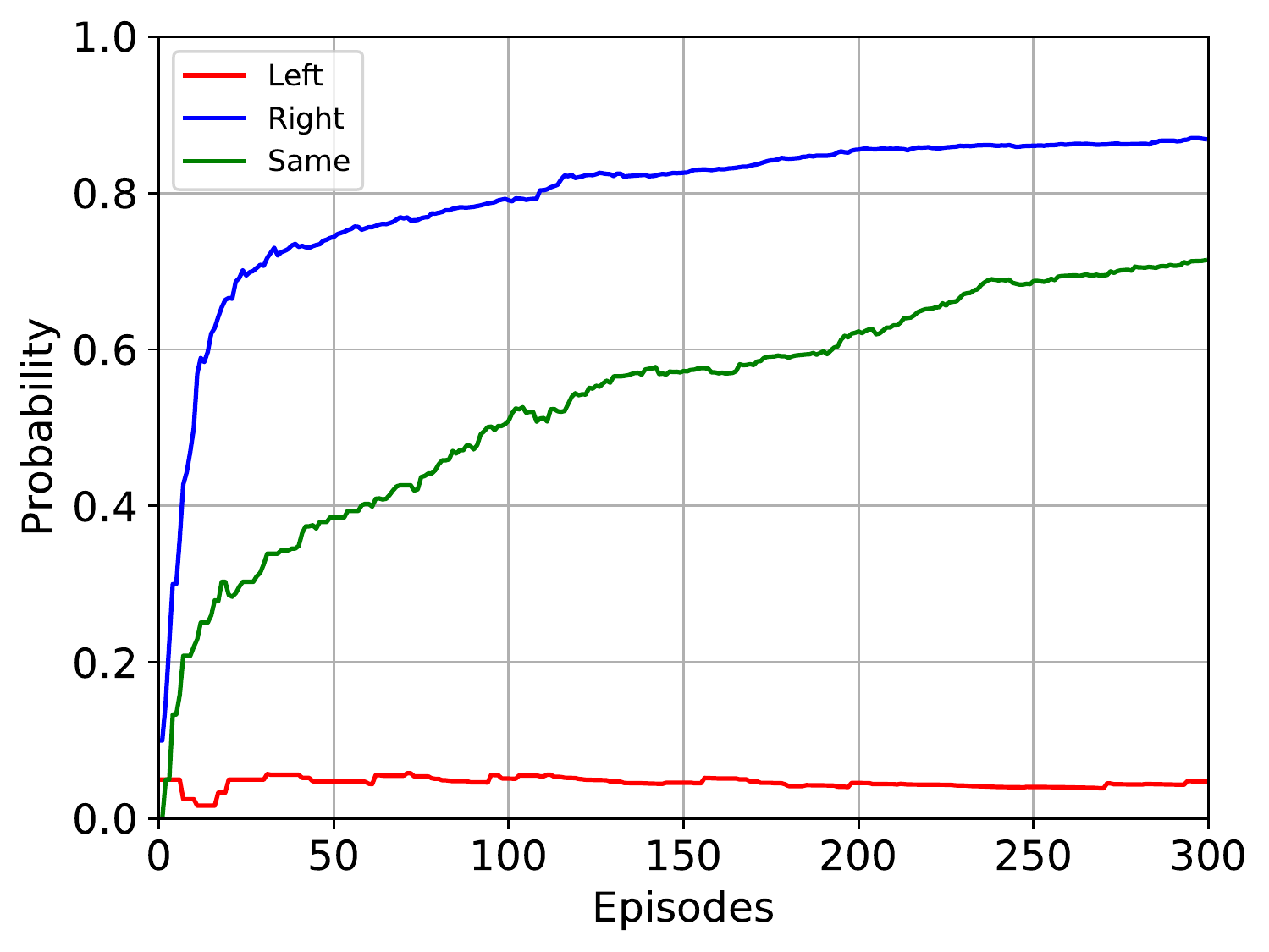} \label{fig:ProbabilityMemoryStochasticS1}}
\\
\subfloat[Learning-based approach.]{\includegraphics[width=0.33\textwidth]{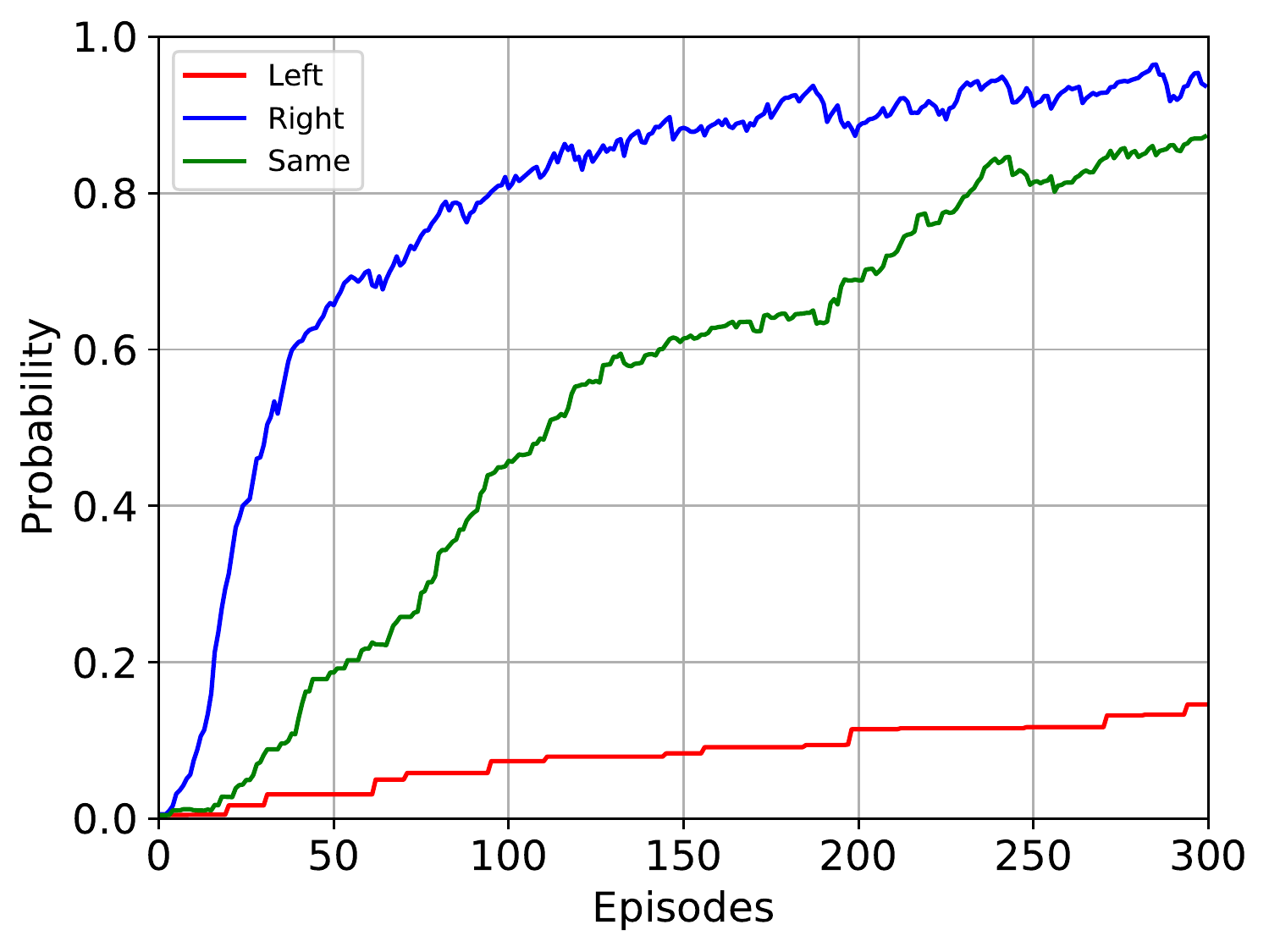} \label{fig:ProbabilityLearningStochasticS1}}
\subfloat[Introspection-based approach.]{\includegraphics[width=0.33\textwidth]{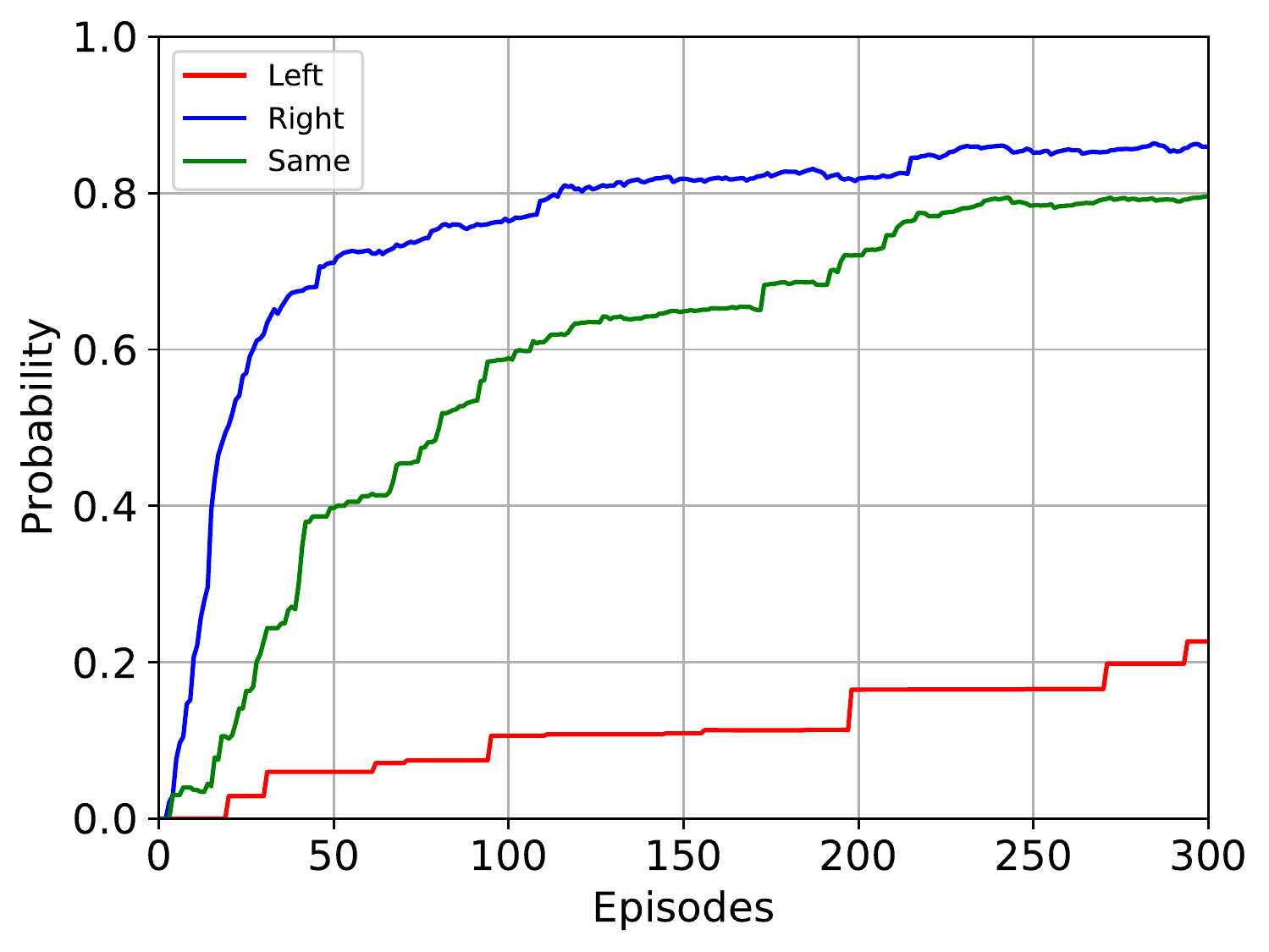} \label{fig:ProbabilityIntrospectionStochasticS1}}
\\
\subfloat{\includegraphics[width=0.35\textwidth]{figures/LabelXRL.pdf}}
\caption{Stochastic robot navigation task. Results are shown from the state $s_1$ for all possible actions. 
  In this case, the agent favors going to the right avoiding stepping out of the level, into the aversive area. 
  In this state, quite different probabilities may be observed for each action, however, all the proposed methods show similar estimated probabilities for the possible actions.
  }
\label{fig:StochasticRobotNavigationS1}
\end{figure*}

\section*{Appendix A: Probabilities of\\success from an internal state}

To have a better understanding of the proposed approaches when using them from internal states, below we show the experimental results for the stochastic robot navigation task from the state $s_1$.
Therefore, the same stochastic scenario and parameters are used as in the previous experiments.
In this case, we only show the results for Q-values, and for the three proposed methods.
Fig.~\ref{fig:QValuesStochasticS1},~\ref{fig:ProbabilityMemoryStochasticS1},~\ref{fig:ProbabilityLearningStochasticS1}, and~\ref{fig:ProbabilityIntrospectionStochasticS1} show the results obtained after 300 learning episodes for the Q-values and the probabilities of success using the memory-based approach, the learning based-approach, and the introspection-based approach, respectively.

When the robot is placed in the state $s_1$, the action of moving to the left $a_L$ leads it to a terminal aversive area, therefore, the associated Q-value and probabilities of success are close to zero.
Nevertheless, due to stochastic transitions, the robot still is able to reach a different room and, hence, the associated Q-value and probabilities of success are not necessarily null.
Conversely, the action of moving to the right $a_R$ has bigger Q-value and probability of success in comparison to staying at the room $a_S$, since $a_R$ leads the robot to the state $s_3$ which is closer to the goal state.
For the obtained probabilities of success using the three approaches, actions of moving right $a_R$ and staying at the same room $a_S$ increase their values in comparison to the Q-value to represent a better estimation of the real probability of success in each case.
The action of moving to the left, in all cases remains low with values near to zero.

For instance, after 300 learning episodes and with the robot just performing an action from $s_1$, an end-user could ask an explanation as: \textit{why did you move to the right in the last situation?}
If the robot uses the probability of success to explain why action $a_R$ has been chosen when being in the state $s_1$, the following explanation could be provided: \textit{`In state $s_1$, I chose to move to the right because it has a probability of success of: (i) $86.91\%$, (ii) $93.75\%$, or (iii) $85.93\%$'}.
The previous values are from using the memory-based, learning-based, and introspection-based approaches respectively.
As previously discussed, we consider the memory-based approach the more accurate estimation for the probability of success, since this value is obtained directly from the actual robot's transitions.
In this case, the Q-value is $0.8166$, therefore, the proposed methods approximate more precisely the real probability and potentially give a more informative answer to a non-expert end-user.
Equivalently, if the end-user asks an explanation as: \textit{why did you not move to the left in the last situation?}
As before, using the probabilities of success to explain the counterfactual of why action $a_L$ has not been chosen when being in the state $s_1$, the explanation can be constructed as follows: \textit{In state $s_1$, I did not choose to move to the left because has only a probability of success of: (i) $4.76\%$, (ii) $14.60\%$, or (iii) $22.67\%$}.
These values also using the memory-based, learning-based, and introspection-based approaches respectively.
The associated Q-value, in this case, is $0.1202$.
Although the deviation is higher, the proposed approaches are able to approximate the probability of success in the stochastic scenario.


\bibliographystyle{ieeetr}
\bibliography{main}

\end{document}